%% file: focal_attention_arr.tex
\documentclass[11pt]{article}
\usepackage[final]{acl}

\usepackage{times}
\usepackage{latexsym}

\usepackage[T1]{fontenc}
\usepackage[utf8]{inputenc}

\usepackage{microtype}

\usepackage{inconsolata}

\usepackage{graphicx}

\input{math_commands.tex}

\usepackage{hyperref}
\usepackage{url}
\usepackage[utf8]{inputenc} % allow utf-8 input
\usepackage[T1]{fontenc}    % use 8-bit T1 fonts
\usepackage{hyperref}       % hyperlinks
\usepackage{url}            % simple URL typesetting
\usepackage{booktabs}       % professional-quality tables
\usepackage{amsfonts}       % blackboard math symbols
\usepackage{nicefrac}       % compact symbols for 1/2, etc.
\usepackage{microtype}      % microtypography
\usepackage{xcolor}         % colors

\usepackage{inconsolata}

\usepackage[english]{babel}
\usepackage{framed}
\usepackage[normalem]{ulem}
\usepackage{algorithm, algorithmic}
\usepackage{amsmath, amsthm, amssymb,amsfonts}
\usepackage{enumerate}
\usepackage{natbib}
\usepackage{xcolor}
\usepackage{comment}
\usepackage{enumitem}

\usepackage{multirow} 
\usepackage{graphicx}
\usepackage{subcaption}
\usepackage{booktabs} % for professional tables
\usepackage{hyperref}
\usepackage{pifont}% http://ctan.org/pkg/pifont
\usepackage{todonotes}
\usepackage{tabularray}
\usepackage{amsmath}
\usepackage{makecell}

\title{Learning to Focus: Focal Attention for Selective and Scalable Transformers}

\author{
  Dhananjay Ram \quad  Wei Xia \quad  Stefano Soatto \\
  AWS AI Labs \\
  \texttt{\{radhna, wxia, soattos\}@amazon.com} \\}

\begin{document}

\maketitle

\begin{abstract}
%The abstract paragraph should be indented \nicefrac{1}{2}~inch (3~picas) on both the left- and right-hand margins. Use 10~point type, with a vertical spacing (leading) of 11~points.  The word \textbf{Abstract} must be centered, bold, and in point size 12. Two line spaces precede the abstract. The abstract must be limited to one paragraph.
%Attention is an integral part of transformer architecture in encoder-only, decoder-only or encoder-decoder models. The softmax attention function used in these models tend to produce noisy probability distribution that can hinder the feature selection at every layer of these models, specially with longer context. In this paper, we propose a novel attention mechanism called Focal Attention to address this problem. We make the attention probability distribution sharper and more selective by using softmax temperature as a hyperparameter or learning it during training. This enables the model to focus on relevant tokens, while ignoring the rest. We show that Focal Attention scales better than the baseline transformer model with model size, training tokens, and context length. Our extensive experiments show that Focal Attention performs significantly better both in terms of loss and downstream task performance requiring 42\% smaller model or 33\% less training data to achieve the same accuracy. The performance gain increases further for a wide variety of long context tasks ranging from 17\% - 82\% of relative improvement.
Attention is a core component of transformer architecture, whether encoder-only, decoder-only, or encoder–decoder model. However, the standard softmax attention often produces noisy probability distribution, which can impair effective feature selection at every layer of these models, particularly for long contexts. We propose \emph{Focal Attention}, a simple yet effective modification that sharpens the attention distribution by controlling the softmax temperature, either as a fixed hyperparameter or as a learnable parameter during training. This sharpening enables the model to concentrate on the most relevant tokens while suppressing irrelevant ones. Empirically, \emph{Focal Attention} scales more favorably than standard transformer with respect to model size, training data, and context length. Across diverse benchmarks, it achieves the same accuracy with up to 42\% fewer parameters or 33\% less training data. On long-context tasks, it delivers substantial relative improvements ranging from 17\% to 82\%, demonstrating its effectiveness in real world applications.
\end{abstract}

\input{intro}
\input{proposal}
\input{exp_setup}
\input{exp_analysis}
\input{related_work}

\section{Conclusion}
We proposed a simple improvement to the attention mechanism used in transformer language models, called \emph{Focal Attention}. This can be implemented very easily with Flash attention for faster training. We show the effectiveness of \emph{Focal Attention} in its ability to scale performance with model size, total training tokens and context length. It performs significantly better than the baseline transformer model for commonsense reasoning tasks. Further evaluation in longer context tasks (up to 64K context length) shows even larger improvement indicating its advantage over base transformer model. We also experimented with adapting a pretrained transformer model with constant temperature scaling, that can yield some improvements, however there exists notable gap with \emph{Focal Attention} model indicating the importance of training from scratch. In future we plan to investigate adapting pretrained models with \emph{Focal Attention} for larger performance improvement. 

\section{Limitations}
The proposed \emph{Focal Attention} showed much improved scaling properties and long context performance. However, it is important to acknowledge its limitations:
\begin{description}[leftmargin=*, nosep]
 \item[Model Architecture] We tested \emph{Focal Attention} with a popular LLaMA-style model architecture. A detailed investigation its effectiveness for other model architectures e.g. Mixture of Experts~\citep{lepikhin2020gshard, fedus2022switch} or models with Multiheaded Latent Attention (MLA)~\citep{liu2024deepseek} is yet to be determined.
 \item[Adapting Pretrained Model] Due to limited compute, we used only 1.3B tokens and adapted baseline transformer to use \emph{Focal Attention} with 5K training steps as discussed in Section~\ref{sec:adapt}. It may be possible to improve the adaptation performance further by fine tuning for longer with more tokens.
% \item[] 
\end{description}

%\subsubsection*{Author Contributions}
%If you'd like to, you may include  a section for author contributions as is done in many journals. This is optional and at the discretion of the authors.

%\subsubsection*{Acknowledgments}
%Use unnumbered third level headings for the acknowledgments. All acknowledgments, including those to funding agencies, go at the end of the paper.

\bibliography{custom}
%\bibliographystyle{plainnat}
%\bibliography{iclr2026_conference}
%\bibliographystyle{acl_natbib}

\appendix
\input{appendix}

\end{document}

%% file: math_commands.tex
\usepackage{amsmath,amsfonts,bm}

\def\eqref#1{equation~\ref{#1}}
\def\1{\bm{1}}

\DeclareMathAlphabet{\mathsfit}{\encodingdefault}{\sfdefault}{m}{sl}
\SetMathAlphabet{\mathsfit}{bold}{\encodingdefault}{\sfdefault}{bx}{n}

%% file: intro.tex
\section{Introduction}
\begin{figure*}
\centering  
 \begin{subfigure}{0.38\linewidth}
\includegraphics[width=\columnwidth]{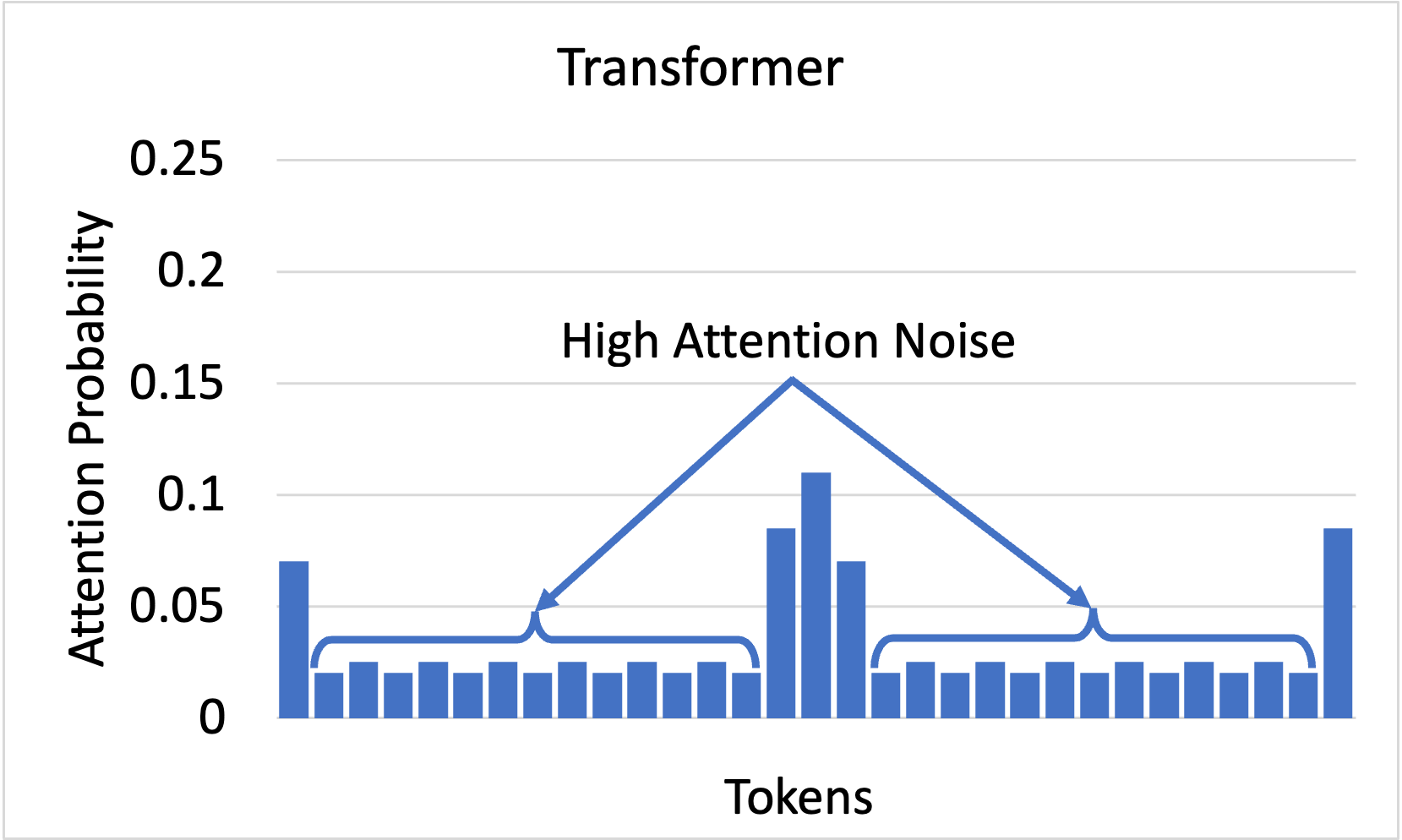}
%\caption{Validation loss w.r.t. model size}
%\label{fig:scale-params}
 \end{subfigure}
 \begin{subfigure}{0.38\linewidth}
 \includegraphics[width=\columnwidth]{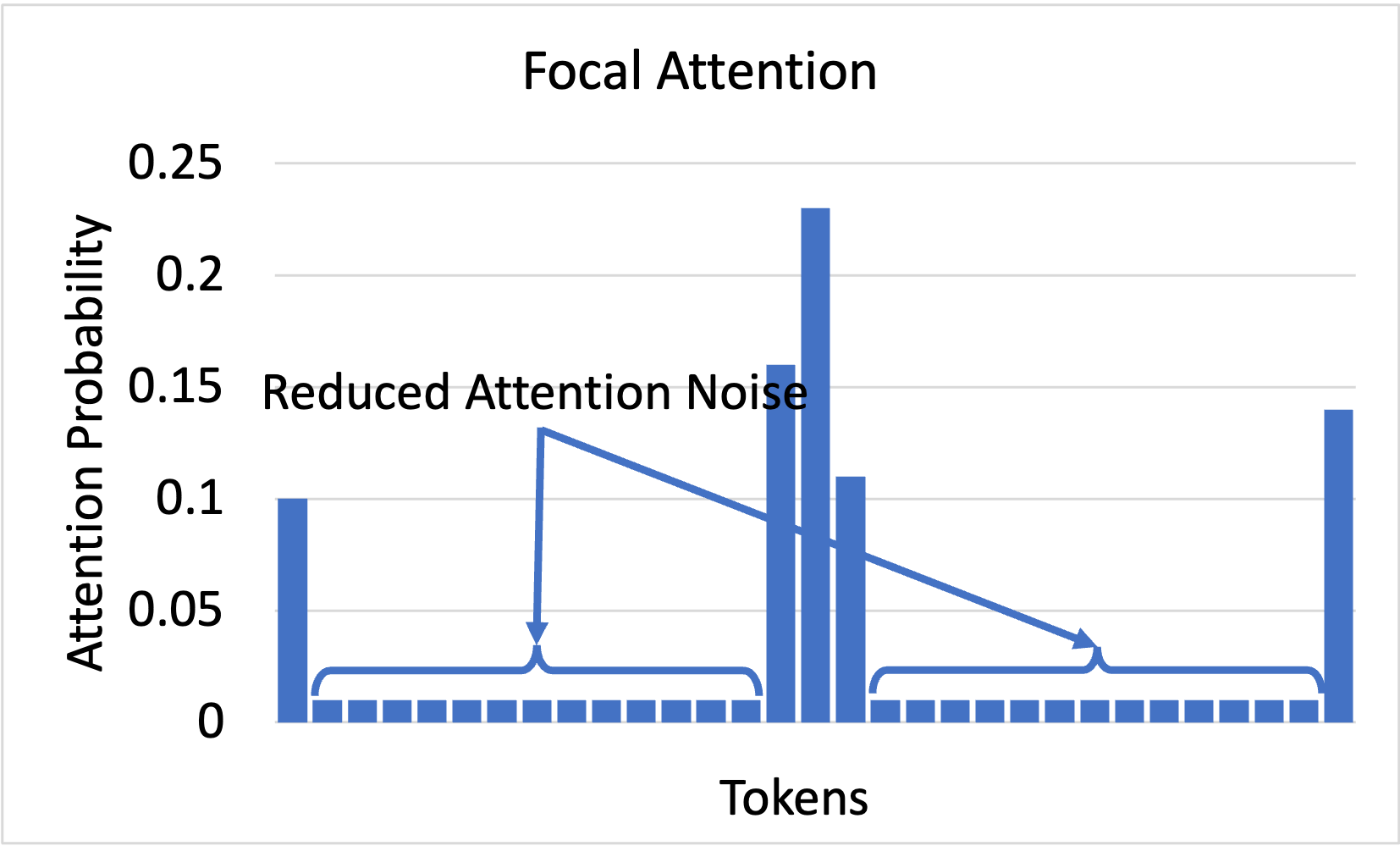}
 %\caption{Downstream task performance w.r.t. model size}
 \end{subfigure}
\caption{Comparison of attention probability distribution of a baseline transformer model and the proposed \emph{Focal Attention}. We see that the \emph{Focal Attention} reduces the attention noise, and redistributes the attention probability from irrelevant tokens to relevant ones.}
 \label{fig:intro}
\end{figure*}

The recent progress in artificial intelligence (AI) has been largely driven by the development of high-quality large language models (LLMs)~\citep{achiam2023gpt, grattafiori2024llama, liu2024deepseek}. These LLMs are trained with transformer architecture~\citep{vaswani2017attention, brown2020language} and attention mechanism lies at the center of transformer model, which enables dynamic weighting of input tokens based on contextual relevance. Attention mechanism computes the importance of each token in the context of a sequence of tokens and is implemented using softmax function. The output of the softmax can be noisy, resulting in assigning some probability mass to even irrelevant tokens as shown in Figure~\ref{fig:intro}. 

%A temperature parameter in softmax can be used to control the sharpness of the probability distribution produced by the softmax function. It adjusts how `confident' the softmax outputs are by scaling the input logits before applying the softmax transformation. This temperature has been used at the output layer of LLMs to produce diverse samples for a given input context to a model~\citep{holtzman2019curious}. It can be very useful generating a diverse set of samples that can be used to improve the model performance at test time~\citep{snell2024scaling}, or can be used fine-tune the model by filtering those samples~\citep{}. However it has not been used in the attention function of the inner layers of transformer to change the sharpness of the distribution and control the feature selection ability at every layer. Generally, a constant temperature equal to input dimension, has been used as proposed in~\citep{vaswani2017attention}. In this work we propose to use a different value for the temperature that can be set as a hyperparameter or learned during model training. Setting a relatively smaller value for the temperature enables the attention to produce sharper output distribution and be more selective for input features. This can be particularly useful for long context tasks as the model can be easily lost in finding relevant information from thousands to hundreds of thousands of tokens~\citep{liu2023lost}. 
A common strategy to denoise the softmax outputs is to introduce a temperature parameter that scales the logits before applying the softmax transformation. Lower temperatures yield more peaked distributions, while higher values result in flatter distributions. This technique has been used at the output layer of LLMs to generate diverse samples during inference~\citep{holtzman2019curious, yue2025does}, and has shown promise in improving performance through selective sampling and test-time inference strategies~\citep{snell2024scaling, muennighoff2025s1}. Despite its utility, temperature scaling has not been explored for attention mechanism in the internal layers of Transformer models.

In the original Transformer~\citep{vaswani2017attention} the attention logits are implicitly scaled by the inverse square root of the input dimensionality~\citep{vaswani2017attention}, which acts as a constant temperature. However, this uniform scaling may not be optimal across all contexts and layers, specially for long-context tasks where identifying relevant information from among thousands (or even hundreds of thousands) of tokens is inherently challenging~\citep{liu2023lost}.

In this work, we explore the role of temperature in attention mechanism and propose \emph{Focal Attention}, where the temperature can be set as a hyperparameter or learned during model training. Setting a relatively smaller value for the temperature, the model sharpens the attention distribution and enables more focused selection of relevant tokens. This can improve feature selection across layers, and enhance the model’s ability to reason over long contexts.
Our contributions can be summarized as follows:
\begin{description}[leftmargin=*, nosep]
    \item[Focal Attention] We propose to explicitly control the softmax temperature within the attention mechanism to generate sharper attention distributions and enable more selective feature extraction (Section~\ref{sec:focal}).
    \item[Improved Performance] We demonstrate that the proposed \emph{Focal Attention} consistently outperforms the standard Transformer model across a range of downstream tasks (Section~\ref{sec:task-perf}).
    %\item We show that the proposed approach performs significantly better than the baseline transformer model on a variety of downstream tasks. 
    \item[Scaling Properties] We show that our approach scales more effectively with model size, training tokens, and context length compared to standard Transformers (Section~\ref{sec:scaling}).
    \item[Long-Context Superiority] \emph{Focal Attention} offers significant gains in long-context scenarios, particularly in settings such as in-context learning, retrieval-augmented generation, and information retrieval, making it highly applicable to real-world applications (Section~\ref{sec:long-context}).
\end{description}

%% file: proposal.tex
% IMPORTANT: talk about simple integration with Flash Attention, enabling faster training

\section{Focal Attention} \label{sec:focal}
We propose \emph{Focal Attention} as an alternative to the attention layer used in transformer models, e.g. GPT~\citep{brown2020language}, LLaMA~\citep{touvron2023llama}, Qwen~\citep{bai2023qwen}, Deepseek~\citep{liu2024deepseek}, etc family of models. This attention is particularly useful for longer context modeling due to its ability to reduce noise in the attention probability and enables better feature selection at every layer. 

The standard softmax function is defined as:
\begin{equation}
    P_i = \frac{exp(z_i)}{\sum_{j=1}^{n}(exp(z_j))}
    %P_i = exp(z_i) / \sum_{j=1}^{n}(exp(z_j))
\end{equation}

When introducing the temperature parameter $t$, the logits are divided by $t$ before applying the softmax:

\begin{equation}
    P_i = \frac{exp(z_i/t)}{\sum_{j=1}^{n}(exp(z_j/t))}
\end{equation}

Here, $z_i$ is the i-th logit (input to softmax) and $t>0$ is the temperature.
The value $t>1$ reduces the magnitude of differences between the logits, making the softmax output more uniform (flatter distribution) and less certain in any single class. This can be useful for encouraging exploration or increasing diversity in sampling (e.g., in text generation). On the other hand, $t<1$ increases the magnitude of differences between the logits, resulting in the softmax output more peaked (concentrated distribution). The highest logit dominates the probability mass, leading to more confident predictions or less diverse sampling.

%\subsection{Effect of Temperature}
%High Temperature ($T>1$) $\implies$ High Entropy

%The logits are divided by a value $T>1$, which reduces the magnitude of differences between them. This makes the softmax output more uniform (flatter distribution). Probabilities are closer together, indicating less certainty or less confidence in any single class. This can be useful for encouraging exploration or increasing diversity in sampling (e.g., in text generation).

%Example: For logits [5, 2, 0] and T=2, the scaled logits become [2.5, 1, 0], resulting in a softer probability distribution. P≈[0.57, 0.28, 0.15]
%Low Temperature ($T<1$) $\implies$ Low Entropy

%The logits are divided by a smaller value $T<1$, which increases the magnitude of differences between them. This makes the softmax output more peaked (concentrated distribution). The highest logit dominates the probabilities, leading to more confident predictions or deterministic sampling.

%Example: For logits [5, 2, 0] and T=0.5, the scaled logits become [10, 4, 0], resulting in a sharper probability distribution,  P≈[0.999, 0.001, 0.000]

\subsection{Transformer with Constant Temperature} \label{sec:focal-const}
Let $X = [X_1, X_2, \hdots, X_n]$ be the input sequence to the transformer model. Here we consider decoder-only architecture as an example to show \emph{Focal Attention}. The model consists of a multi-layer architecture where each layer has an attention module followed by a Feed-forward module~\citep{vaswani2017attention, touvron2023llama}. Our proposed \emph{Focal Attention} updates the attention module, keeping other parts of the model same. Any given input $X$ to the attention module is projected into the query ($Q$), key ($K$) and value ($V$) space. The Query and Key representations are used to compute the attention probability, which is multiplied with the Value representation to yield the output.
\begin{equation}
    Q = XW^Q, K = XW^K , V = XW^V
\end{equation}
\begin{equation}
    Attention(X) = softmax{(\frac{QK^T}{\sqrt{d}})} V    
\end{equation}
where $W^Q$, $W^K$, $W^V$ are parameter matrices. The softmax temperature is $\sqrt{d}$ and it is shown to produce noisy probability distribution~\cite{}, as shown in Figure~\ref{fig:intro}. 
In this work, we propose to reduce the temperature by using a scaling factor $t < 1$ as a hyperparameter to the model:
\begin{equation} \label{eq:focal-const}
    Attention(X) = softmax{(\frac{QK^T}{t \sqrt{d}})} V    
\end{equation}
This produces a sharper probability distribution enabling the model to better focus on relevant tokens and ignore the irrelevant ones. We can use a different scaling factor per layer, however we use a constant scaling for the whole model.

%The attention probability distribution for transformer model is computed as:
%\begin{equation}
%    P_i = exp(z_j/\sqrt{d}) / \sum_{j}(exp(z_j/\sqrt{d}))
%\end{equation}
%where $z_i$ is the input logit and $d$ is the input dimension. We propose to learn a hyperparameter $t$ to scale the divisor and adjust the scaling factor to learn a better attention probability distribution for our problem as follows:
%\begin{equation}
%    P_i = exp(z_j/t\sqrt{d}) / \sum_{j}(exp(z_j/t\sqrt{d}))
%\end{equation}

\subsection{Transformer with Learned Temperature} \label{sec:focal-learn}
We propose to further generalize the previous approach by learning the temperature ($\tau$) from the hidden representation. This enables the model to learn different scaling factors for different layers in the model, which is desirable because lower layers may require softer attention probability distribution and higher layers may require sharper attention probability distribution as the model get more confident. More specifically, we learn the softmax temperature using the input to the attention module ($X \in \mathbb{R}^{n \times d}$) of the model by learning a parameters vector ($w_{\tau} \in \mathbb{R}^d$) of the size of hidden states in every layer as shown in Eq~\ref{eq:learn-temp}
\begin{equation} \label{eq:learn-temp}
    \tau = clip(mean(X w_{\tau}), \tau_{min}, \tau_{max})
\end{equation}
We clip the temperature value to a predefined minimum ($\tau_{min}$) and maximum ($\tau_{max}$) value as the temperature tends to drift much further from the baseline leading to performance degradation (Section~\ref{sec:effect-learned}). The resulting attention output is presented in Eq~\ref{eq:focal-learn} 
\begin{equation} \label{eq:focal-learn}
    Attention(X) = softmax{(\frac{QK^T}{\tau})} V    
\end{equation}

%% file: exp_setup.tex
\begin{figure*}
\centering  
 \begin{subfigure}{0.38\linewidth}
\includegraphics[width=\columnwidth]{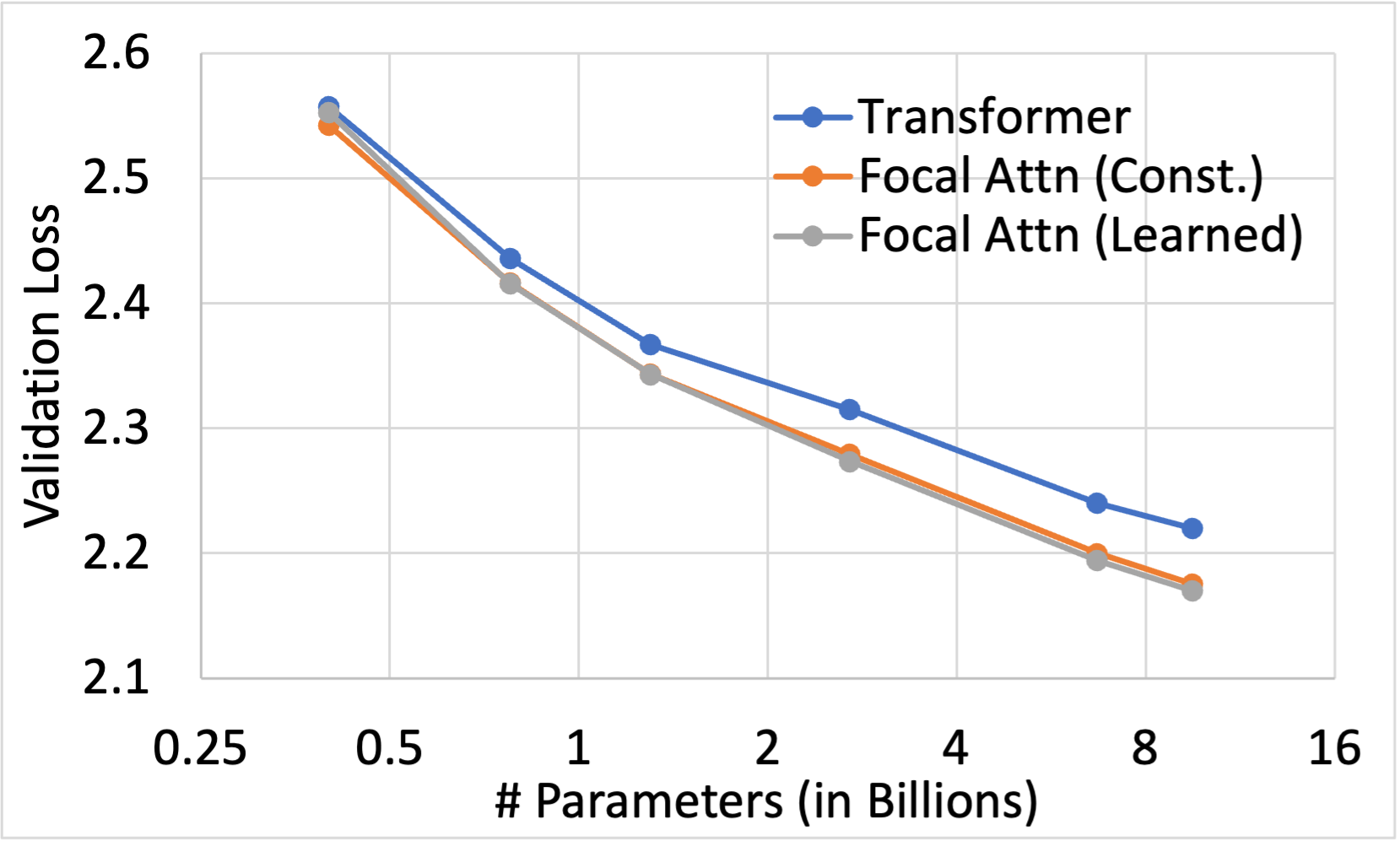}
\caption{Validation loss w.r.t. model size}
\label{fig:scale-params-loss}
 \end{subfigure}
 \begin{subfigure}{0.38\linewidth}
 \includegraphics[width=\columnwidth]{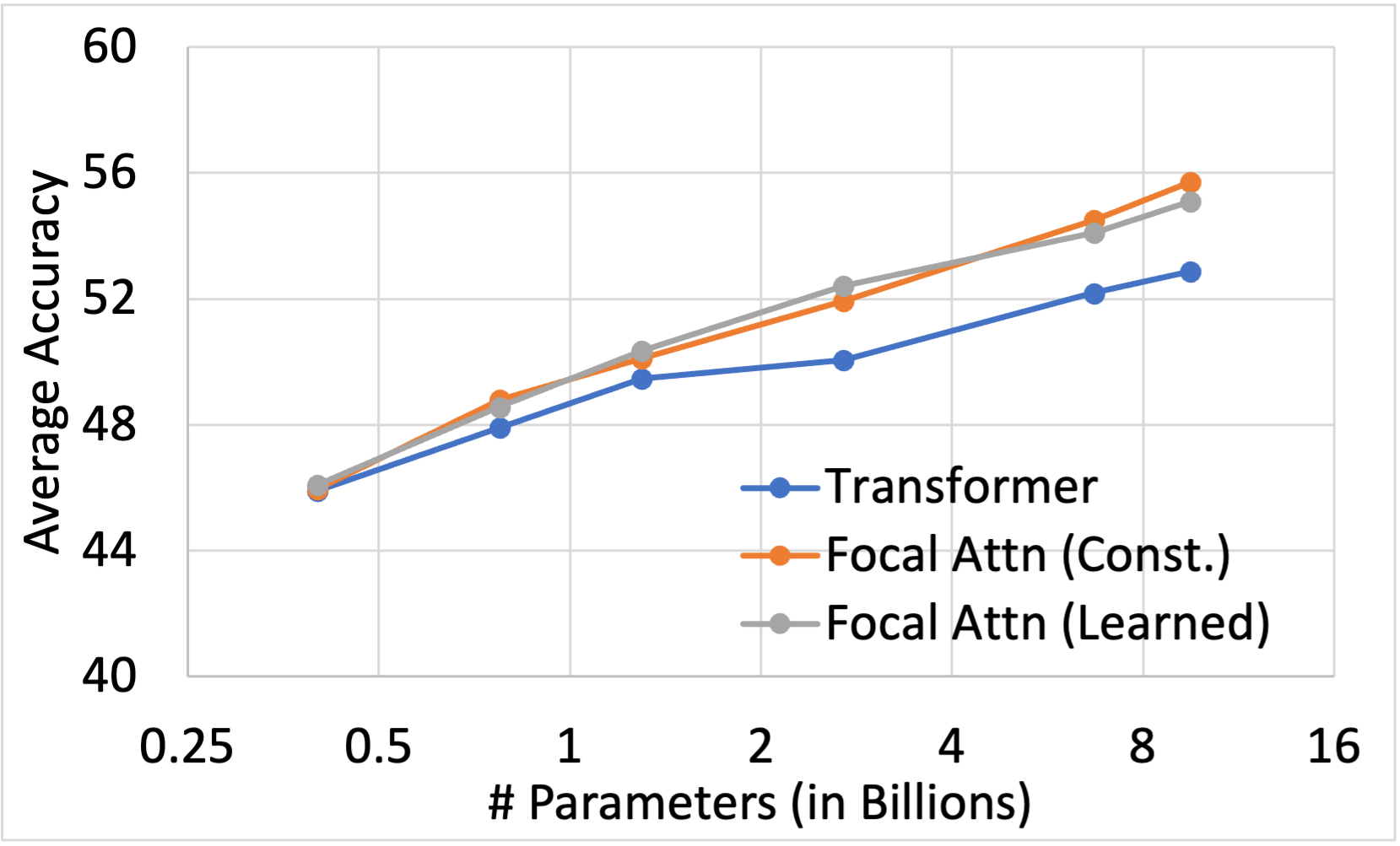}
 \caption{Task performance w.r.t. model size}
 \label{fig:scale-params-task}
 \end{subfigure}
\caption{Scaling model size from 400M to 9.5B: \emph{Focal Attention} scales better than the base transformer model with model size. The performance gap increases with larger models.}
\end{figure*}

\begin{figure*}
\centering  
 \begin{subfigure}{0.38\linewidth}
\includegraphics[width=\columnwidth]{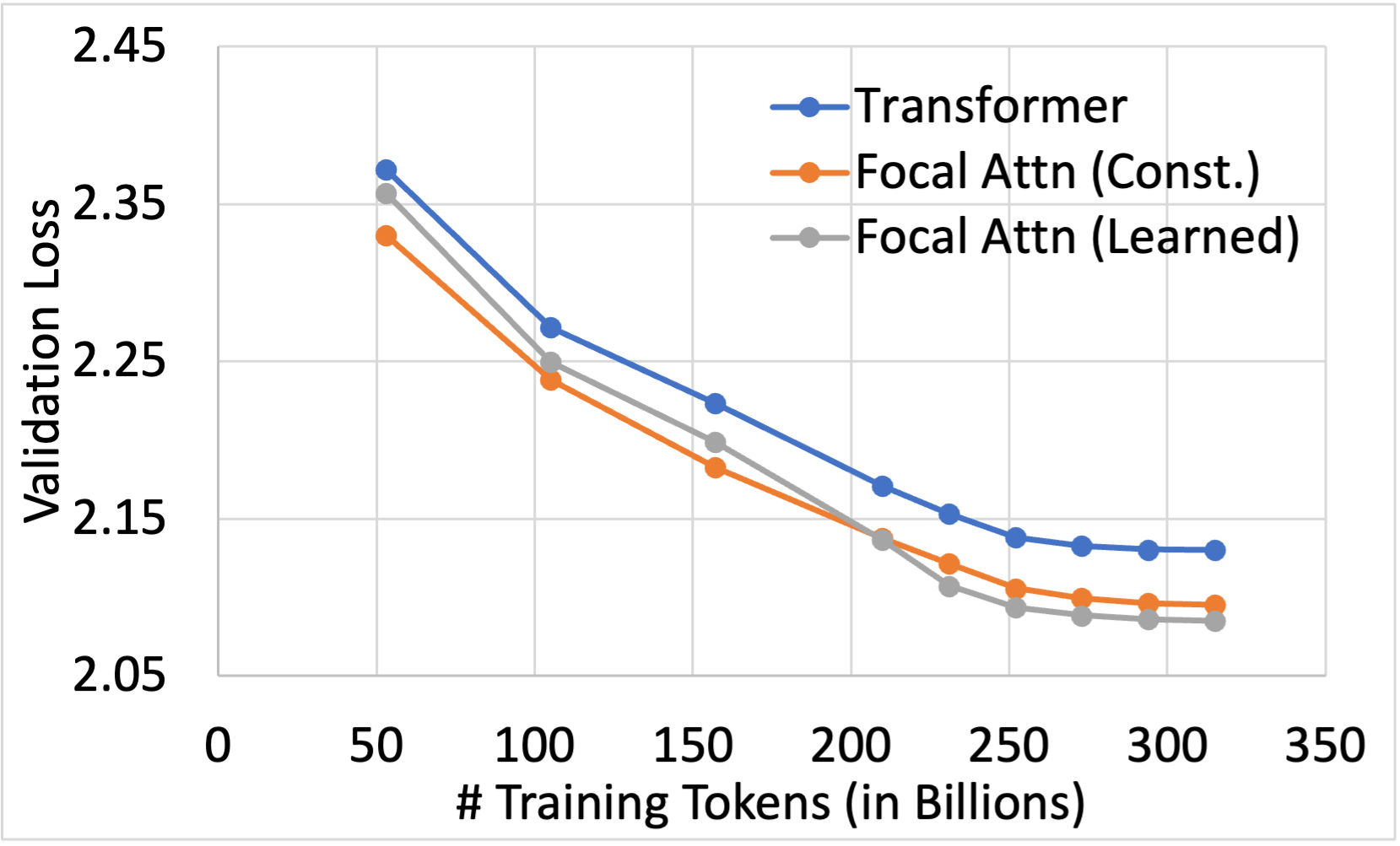}
\caption{Validation loss w.r.t. training tokens}
\label{fig:scale-tokens-loss}
 \end{subfigure}
%\hfill
 \begin{subfigure}{0.38\linewidth}
 \includegraphics[width=\columnwidth]{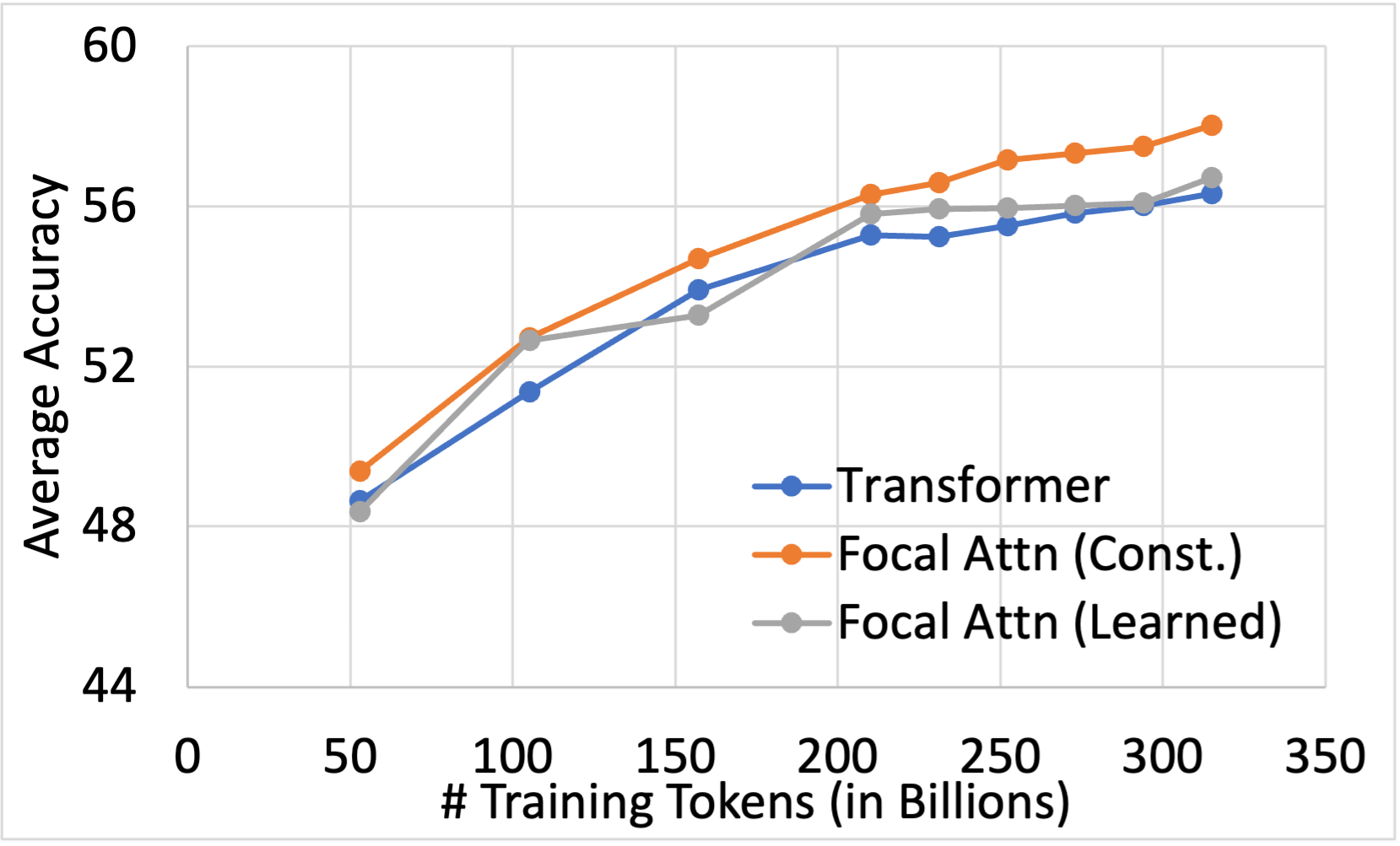}
 \caption{Task performance w.r.t. training tokens}
 \label{fig:scale-tokens-task}
 \end{subfigure}
\caption{Scaling total training tokens to 315B: \emph{Focal Attention} scales better than the base transformer model with increasing amount of training data. The performance gain is larger with more training tokens used.}
\end{figure*}

\section{Experimental Setup}
\subsection{Model Architecture}
We use LLaMA style model~\citep{touvron2023llama} architecture for our experiments to show the effectiveness of the proposed \emph{Focal Attention} approach. This is a decoder-only language model with SwiGLU activations~\citep{shazeer2020glu}, RMS norm~\citep{zhang2019root} for pre-normilaztion with rotary positional embedding~\citep{su2024roformer}. We also use query-key normalization~\citep{henry2020query} for stable training and improved performance, providing us a strong baseline model. We use 6 different sized models ranging from 400M to 9.5B in total parameters for our experiments (details in appendix~\ref{sec:apx-arch}). 

\subsection{Dataset and Training} \label{sec:train}
We train our models using a filtered mixture of publicly available large scale pretraining datasets similar to Open-LLaMA~\citep{openlm2023openllama}. We utilize LLaMA tokenizer~\citep{touvron2023llama} with online sample packing~\citep{fan2024hlat} to tokenize and efficiently pack the tokens during training, instead of doing them separately.

%\subsubsection{Pretraining}\label{sec:pretrain}
We train 6 different sized models from scratch: 400M, 777M, 1.3B, 2.7B, 6.7B and 9.5B. For each model size, we train 3 variants: baseline transformer, \emph{Focal Attention} with constant temperature, and \emph{Focal Attention} with learned temperature. These models were trained with the same batch size of 0.26M tokens, for 100K steps for a total of 26B tokens.
We use AdamW optimizer~\citep{loshchilov2017decoupled} for all models with $\beta_1 = 0.9$, $\beta_2 = 0.95$, weight decay of $0.05$, gradient clipping of $1$ and warmup of 2000 steps. The models are trained with a cosine learning rate schedule where the learning rate is decayed to 10\% of the peak learning rate. We train the models with bf16 precision, and use flash attention~\citep{dao2022flashattention} for faster training (details in appendix~\ref{sec:apx-hyp}). 

\subsection{Training with Large Scale Data} \label{sec:token-scale}
Additionally, we train a 2.7B model for 300B tokens from scratch, to evaluate \emph{Focal Attention} for large amount of training token scenario. The model is trained with a batch size of 2.1M tokens and learning rate of $5\times10^{-4}$ for 150K steps or 315B tokens, with the same optimizer setting as in previous section. 

\subsection{Long Context Fine-Tuning}
To measure the effectiveness of \emph{Focal Attention} at longer context, we consider the 2.7B model trained for 315B tokens with context length 2048. We adapt both models (base transformer and \emph{Focal Attention}) for 32768 context length by fine tuning it for an additional 5B tokens at that length with longer training samples. We scale up the RoPE $\theta$ from 10K to 500K for this purpose following~\citep{xiong2023effective}. 

\subsection{Evaluation Framework}
We evaluate all models for commonsense reasoning tasks, and longer context models on various tasks designed for longer context as discussed in the following sections.
\subsubsection{Commonsense Reasoning}
%We evaluated the Commonsense Reasoning ability of the models using LM-Evaluation-Harness~\citep{eval-harness} framework. More specifically, we use ARC-Easy - Multiple-choice science questions from grade-school exams designed to test basic factual and commonsense reasoning, ARC-Challenge, BoolQ, Hellaswag, Lambada, PIQA and Winogrande.  
We evaluate the models on a diverse set of language understanding benchmarks using LM-Evaluation-Harness~\citep{eval-harness} framework. The tasks include ARC-Easy and ARC-Challenge~\citep{clark2018think} for science question answering at varying difficulty levels, BoolQ~\citep{clark2019boolq} for binary QA over Wikipedia passages, HellaSwag~\citep{zellers2019hellaswag} for commonsense inference, LAMBADA~\citep{paperno2016lambada} for long-range word prediction, PIQA~\citep{bisk2020piqa} for physical reasoning in everyday scenarios, and Winogrande~\citep{sakaguchi2021winogrande} for coreference resolution requiring commonsense knowledge.

\subsubsection{Long Context Evaluation} \label{sec:eval-long}
We evaluate the long context capability of our models with HELMET~\citep{yen2025helmet} framework. The tasks are categorized in 6 groups, with each group consisting of multiple datasets: (a) Many-shot in-context learning (ICL), (b) Long Document Question Answering (LongQA), (c) Retrieval-Augmented Generation (RAG), (d) Passage Reranking, (e) Generation with Citations, and (f) Synthetic Recall (details in Appendix~\ref{sec:apx-eval-long}).

%% file: exp_analysis.tex
\section{Experimental Analysis}
In this section, we first analyze the scaling properties of the proposed \emph{Focal Attention} in different dimensions followed by comparing its commonsense reasoning performance with baseline transformer model. Finally, we measure its ability to perform tasks at much longer context to show its usefulness in real world scenarios. 

\subsection{Scaling Properties} \label{sec:scaling}
We study the scaling properties of \emph{Focal Attention} in three dimensions: (i) model size, (ii) total training data and (iii) context length as discussed in the following sections.

\begin{figure*}
\centering  
 \begin{subfigure}{0.32\linewidth}
\includegraphics[width=\columnwidth]{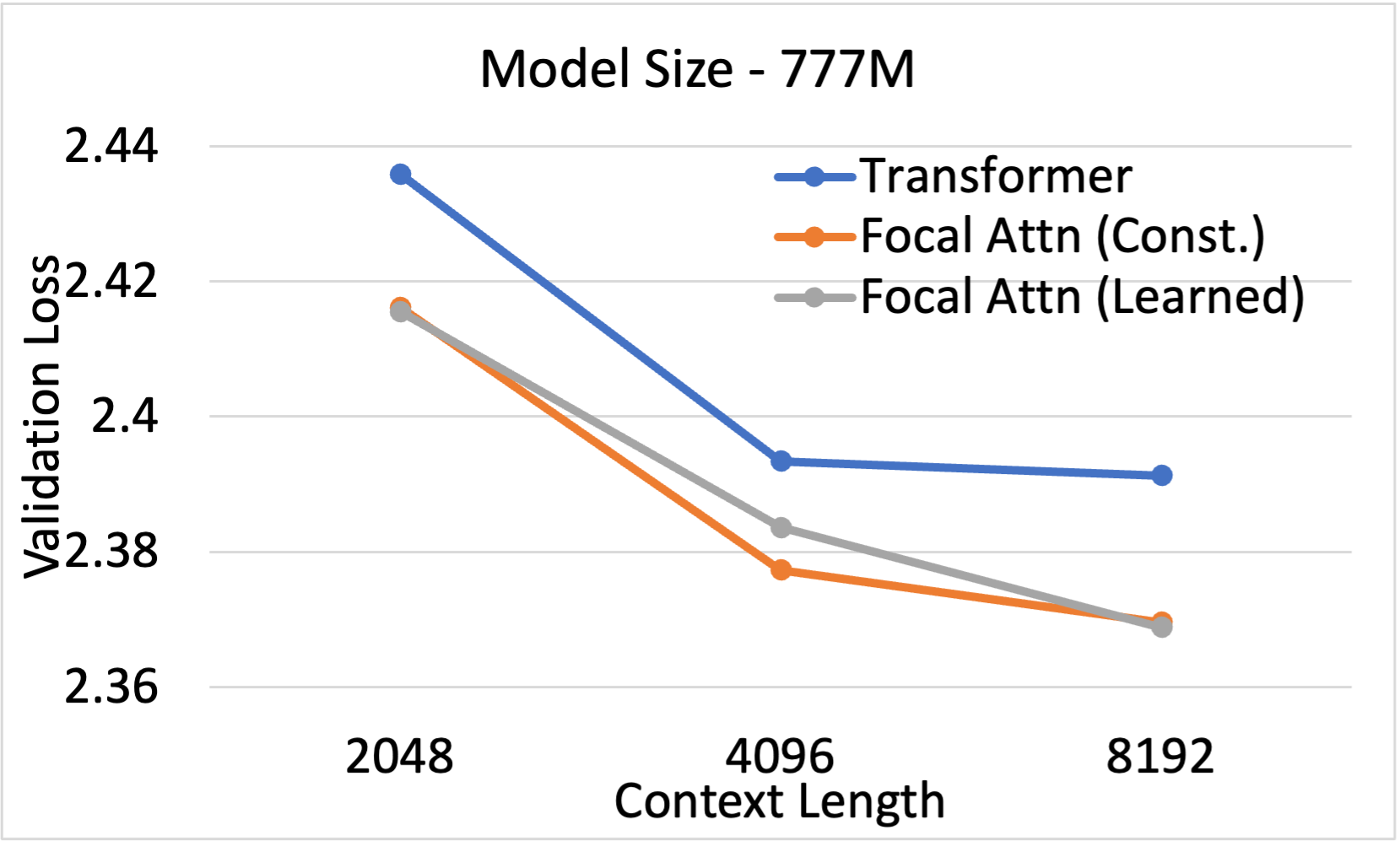}
%\caption{Scaling model size from 400M to 9.5B}
 \end{subfigure}
%\label{fig:dem}
 \begin{subfigure}{0.32\linewidth}
 \includegraphics[width=\columnwidth]{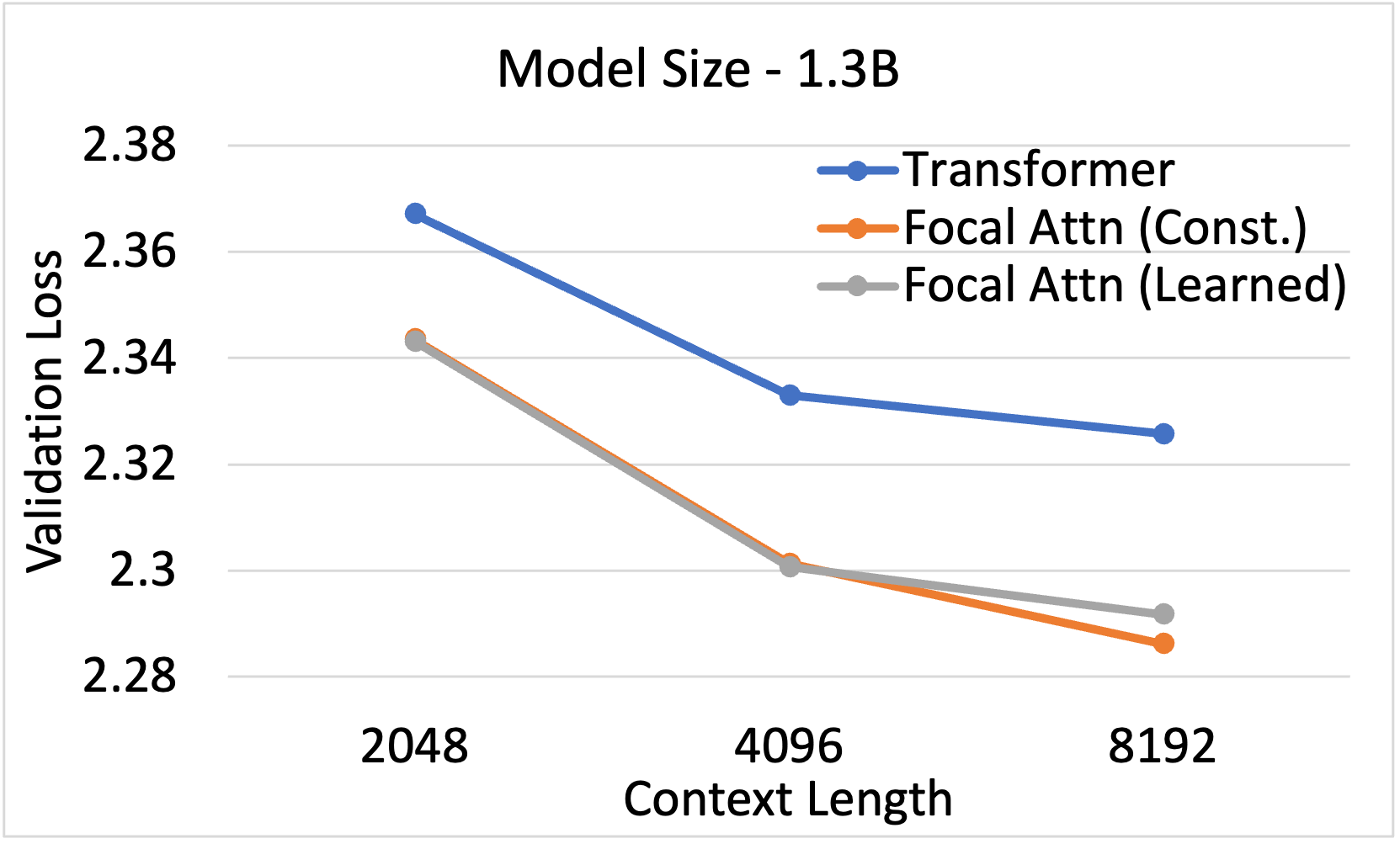}
 %\caption{Scaling training tokens to 300B for 2.7B model}
 \end{subfigure}
 \begin{subfigure}{0.32\linewidth}
 \includegraphics[width=\columnwidth]{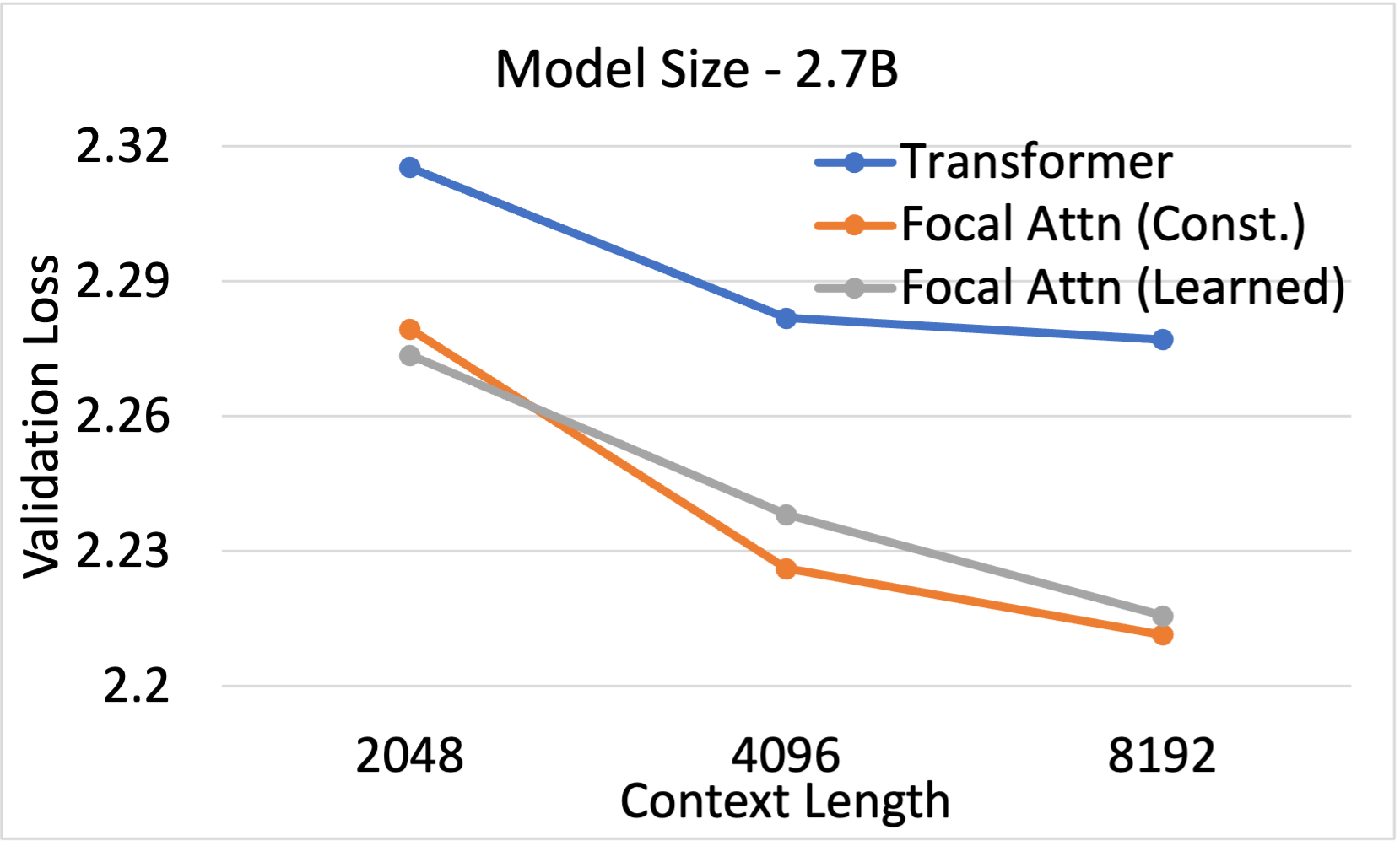}
 %\caption{Scaling training tokens to 300B for 2.7B model}
 \end{subfigure}
\caption{Context length expansion: Training from scratch of 3 different model sizes. \emph{Focal Attention} improves performance for all model sizes, the improvement is larger for longer context length.}
\label{fig:scale-context}
\end{figure*} 

\begin{table*}
\centering
\resizebox{0.9\textwidth}{!}{%
\setlength{\tabcolsep}{3pt}
\begin{tabular}{l|cccccccc} 
\toprule
 \textbf{Model} & \textbf{ARC-C} & \textbf{ARC-E} & \textbf{BoolQ} & \textbf{Hellaswag} & \textbf{LAMBADA}  & \textbf{PIQA} & \textbf{Winogrande} & \textbf{Average}\\ 
\midrule 
Baseline & 30.2 & 57.49 & 62.32 & 59.2 & 51.06 & 73.88 & 56.67 & 55.83 \\
\midrule 
%Open-LLaMA-3B-v2 & 36 & 63 & 66 & 70 & 58 & 78 & 63 & 62 \\
%\midrule
Focal Attn (Constant) & {\bf 32.08} & {\bf 60.52} & {\bf 63.79} & {\bf 62.12} & 53.85 & {\bf 74.16} & {\bf 59.59} & {\bf 58.02} \\
Focal Attn (Learned) & 31.66 & 56.4 & 62.63 & 57.38 & {\bf 57.85} & 73.18 & 57.85 & 56.71 \\
\bottomrule
\end{tabular}
}
\caption{\emph{Focal Attention} model performance compared to the baseline transformer, trained for 315B tokens.} 
\label{tab:task-perf}
\end{table*}

\subsubsection{Scaling Model Size}
We trained 6 models of different sizes ranging from 400M to 9.5B, with context length of 2048 to analyze the scaling performance of \emph{Focal Attention} with number of parameters, as discussed in Section~\ref{sec:train}. 
%These models were trained with the same batch size of 256K tokens, for 100K steps for a total of 25B tokens. The detailed hyperparameters are presented in Appendix. The final validation loss and average accuracy (on LM-Evaluation-Harness) for these models are shown in Figures~\ref{fig:scale-params-loss} and\ref{fig:scale-params-task} respectively.
We evaluate these models on LM-Evaluation-Harness tasks and plot the average accuracy with respect to model size in Figure~\ref{fig:scale-params-task}. The corresponding validation losses are shown in Figures~\ref{fig:scale-params-loss}. Both variants of \emph{Focal Attention} perform significantly better than the base transformer for all model sizes, where the performance improvement increases with model size. We observe that \emph{Focal Attention} can achieve similar performance as baseline with 42\% less parameters. 

\subsubsection{Scaling Training Data} \label{sec:scale-data}
We analyze how \emph{Focal Attention} scales with large scale training data by training the 2.7B model for 315B tokens from scratch as discussed in Section~\ref{sec:token-scale}. We compare the validation loss and average accuracy (on LM-Evaluation-Harness) of \emph{Focal Attention} with base transformer model at different amount of training tokens (intermediate checkpoints) in Figures~\ref{fig:scale-tokens-loss} and~\ref{fig:scale-tokens-task} respectively. We observe that the loss and downstream task accuracy is significantly better for both variants of \emph{Focal Attention} for all checkpoints and the improvement is larger at the later stage of training, indicating a positive trend for training with even larger datasets. More specifically, the checkpoint at 210B tokens for \emph{Focal Attention} (Const.) yields similar performance as the final checkpoint of baseline transformer model showing that \emph{Focal Attention} requires 33\% less training data to achieve similar performance as baseline.

\subsubsection{Scaling Context Length}
We evaluate the ability of \emph{Focal Attention} to scale performance with context length by training 3 different sized models (777M, 1.3B and 2.7B) with 3 context lengths (2048, 4096 and 8192) from scratch, using the same hyperparamaters as discussed in Section~\ref{sec:train}. While changing the context length, we keep the total batch equal to 0.26M tokens for all models. We present the validation loss for each model for all three context length in Figure~\ref{fig:scale-context} and observe significant performance improvement by using \emph{Focal Attention}. Both variants of \emph{Focal Attention} perform similarly for all models and larger improvement is observed for longer context length. This shows that the \emph{Focal Attention} scales better than the base transformer model at various context lengths.

\subsection{Downstream Task Performance} \label{sec:task-perf}
We compare the performance of 2.7B \emph{Focal Attention} models (trained for 315B tokens) with baseline transformer on commonsense reasoning tasks in LM-Evaluation-Harness. The task specific and average performance is presented in Table~\ref{tab:task-perf}. We observe that both \emph{Focal Attention} models perform better than the baseline transformer with an average of 2.2 points absolute improvement, for \emph{constant temperature scaling}. The \emph{learned temperature} variant performs worse than using \emph{constant temperature}, except for LAMBADA task. Due to this limited performance of \emph{learned temperature} variant, we only consider \emph{constant temperature scaling} for evaluating long context capability of \emph{Focal Attention}, as discussed in the following section. 

\begin{figure*}
\centering  
 \begin{subfigure}{0.3\linewidth}
 \includegraphics[width=\columnwidth]{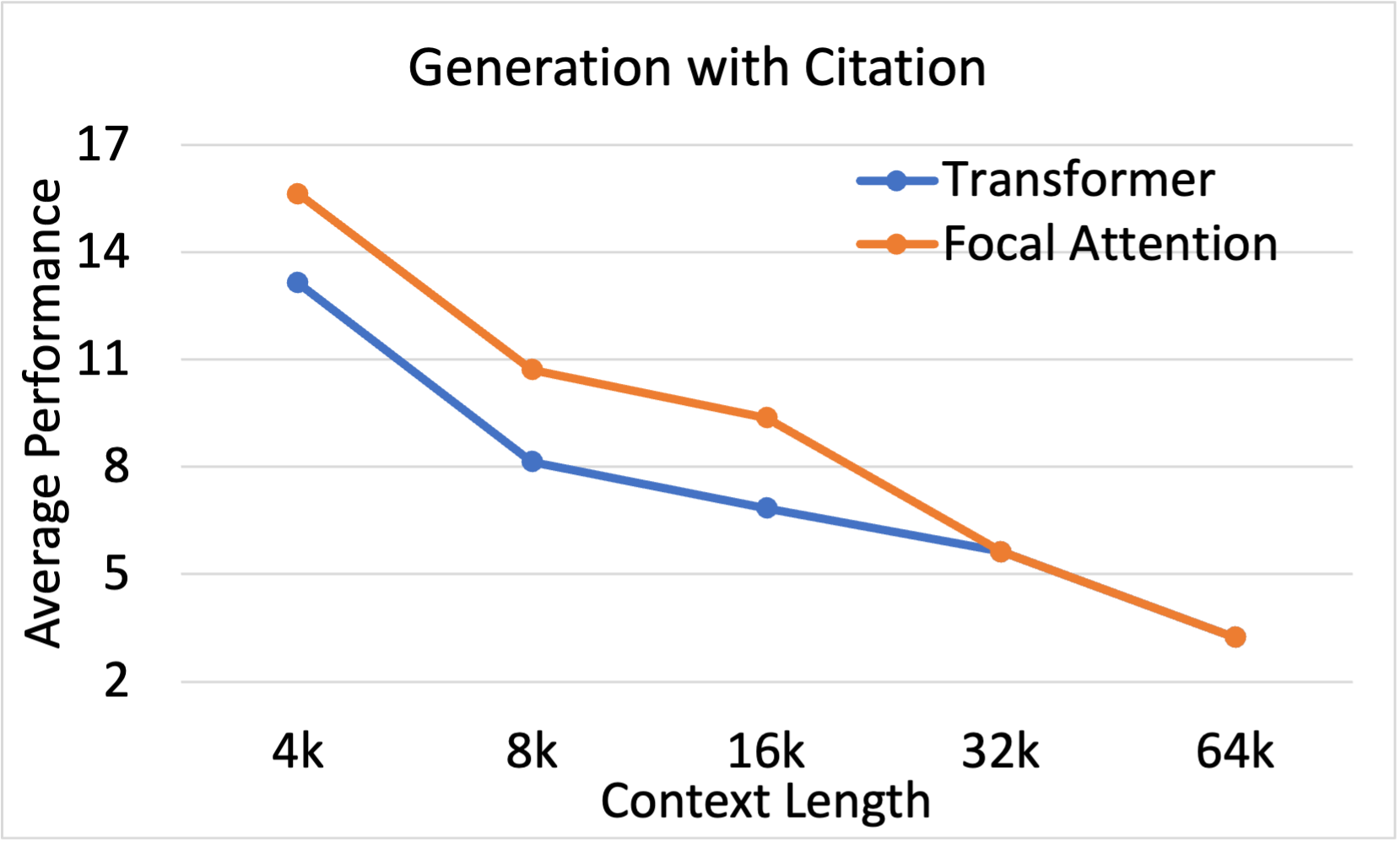}
 \end{subfigure}
 \begin{subfigure}{0.3\linewidth}
 \includegraphics[width=\columnwidth]{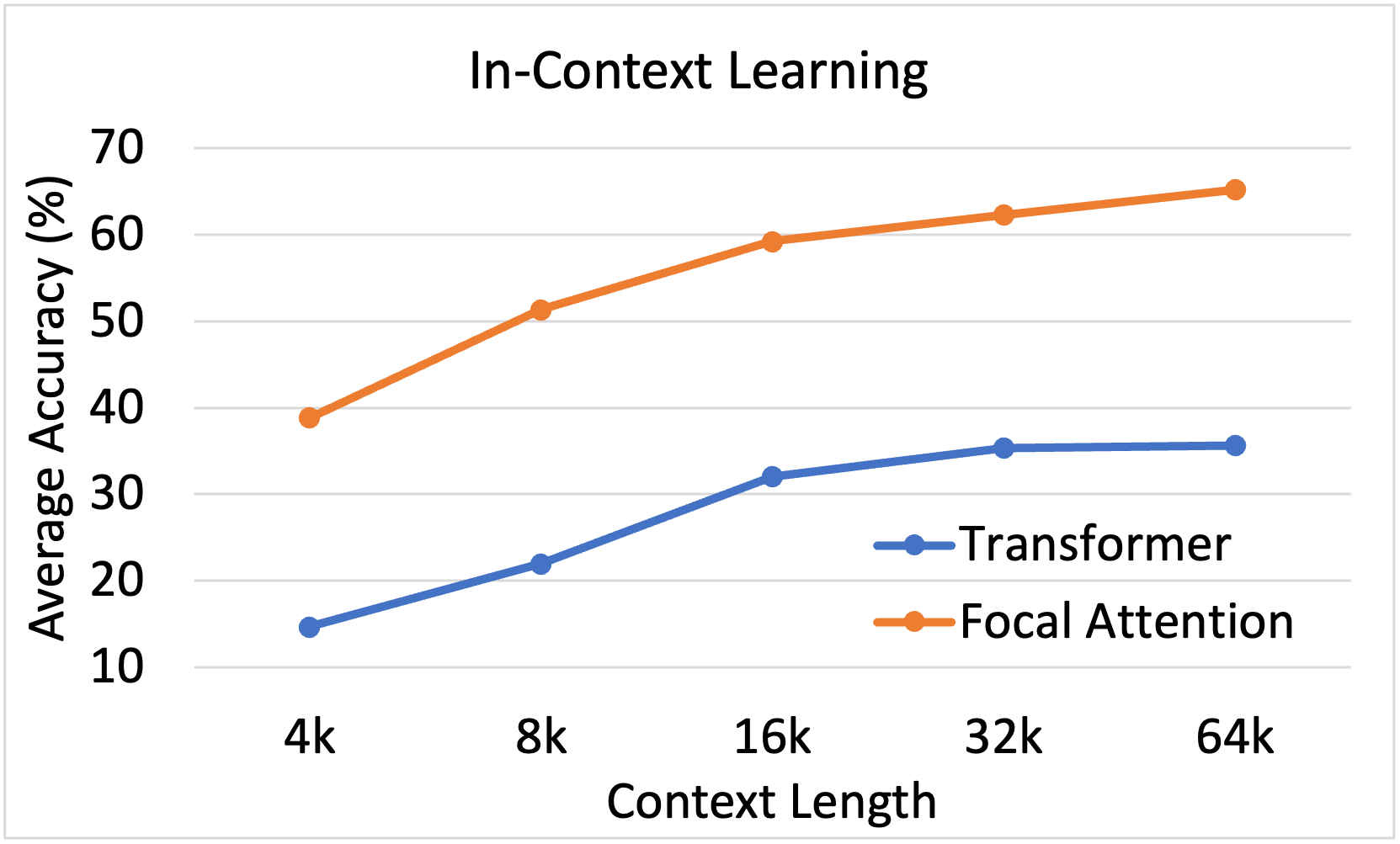}
 \end{subfigure}
 \begin{subfigure}{0.3\linewidth}
 \includegraphics[width=\columnwidth]{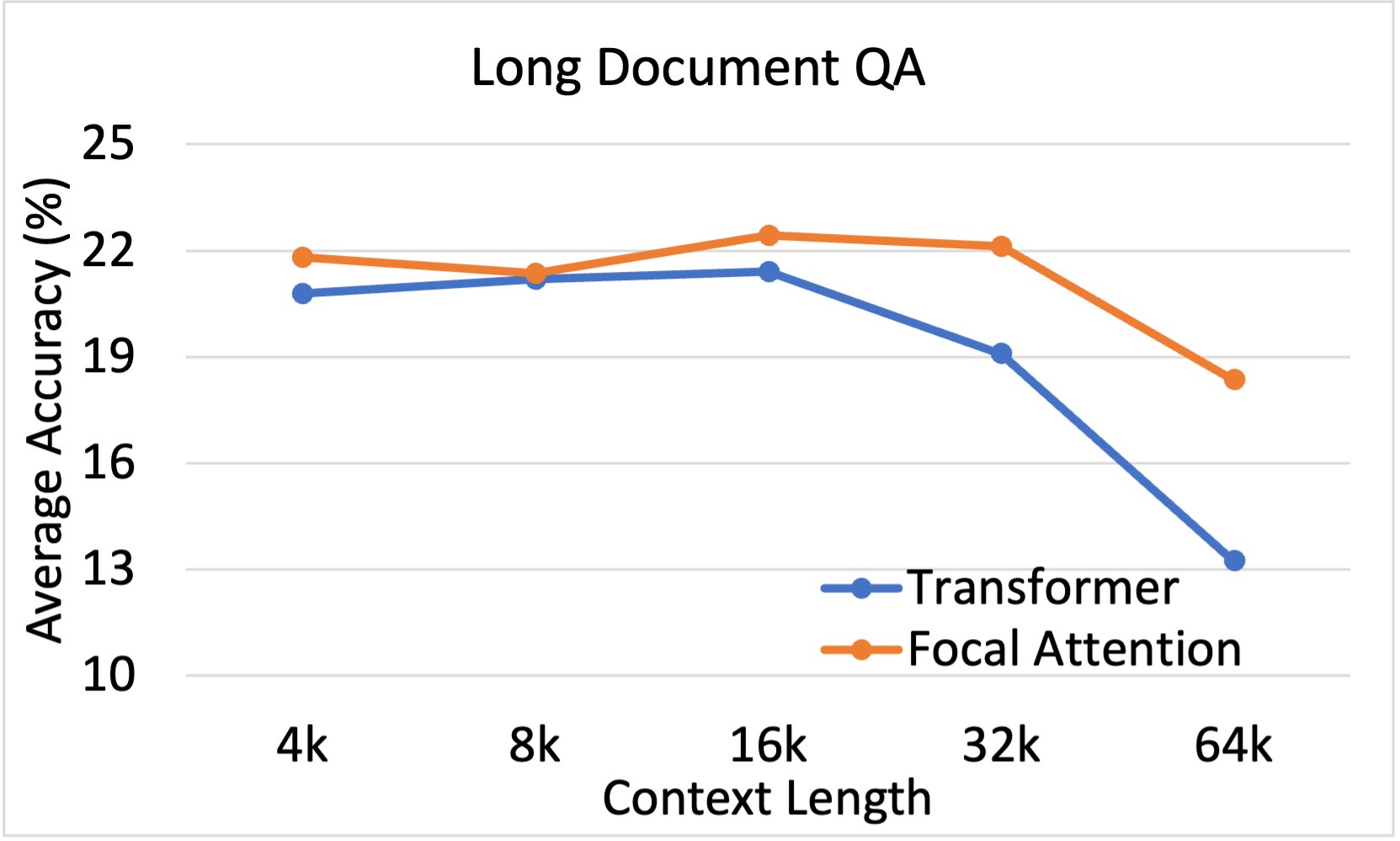}
 \end{subfigure}
%\vspace{1cm}
\centering  
 \begin{subfigure}{0.3\linewidth}
 \includegraphics[width=\columnwidth]{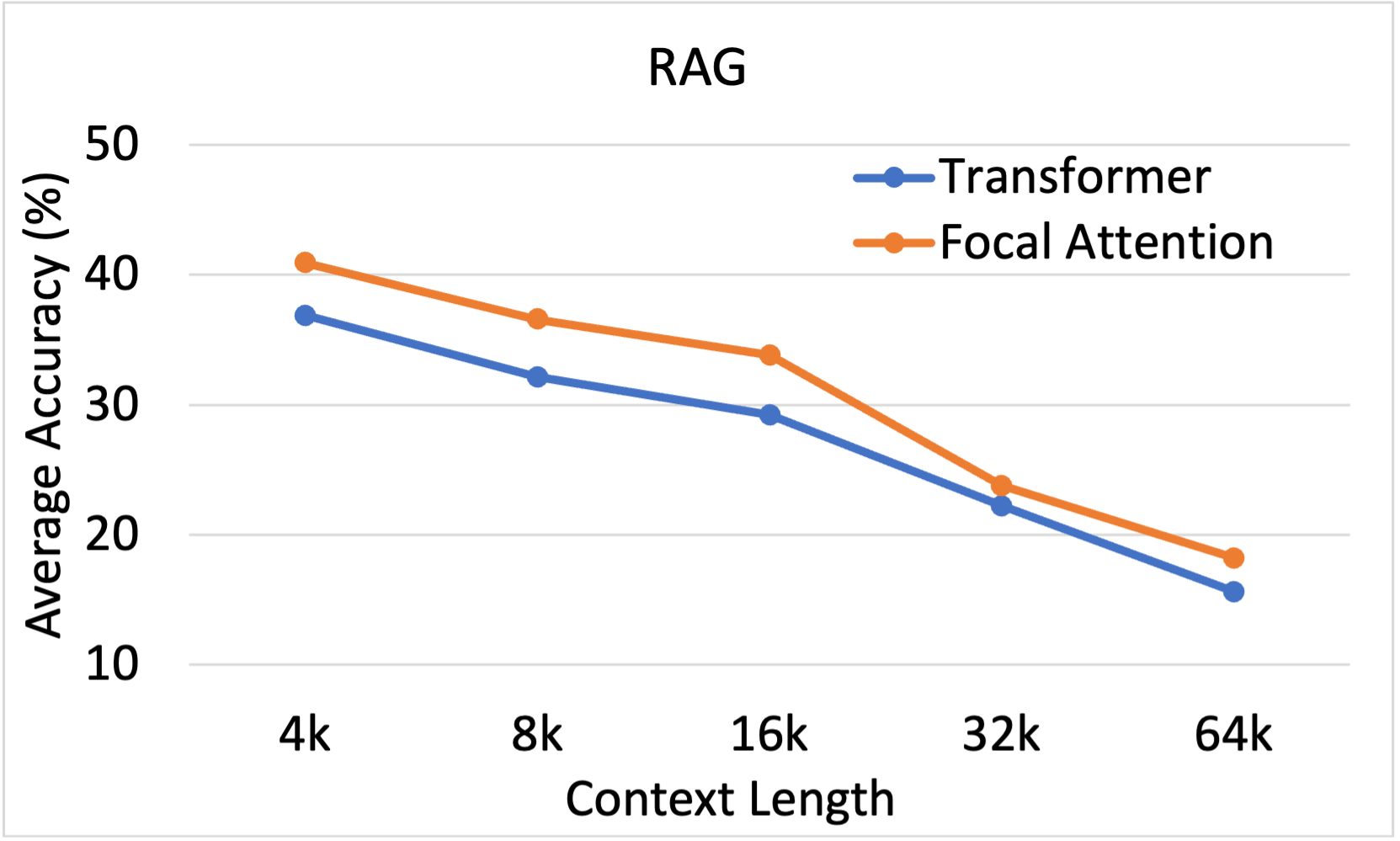}
 \end{subfigure}
 \begin{subfigure}{0.3\linewidth}
 \includegraphics[width=\columnwidth]{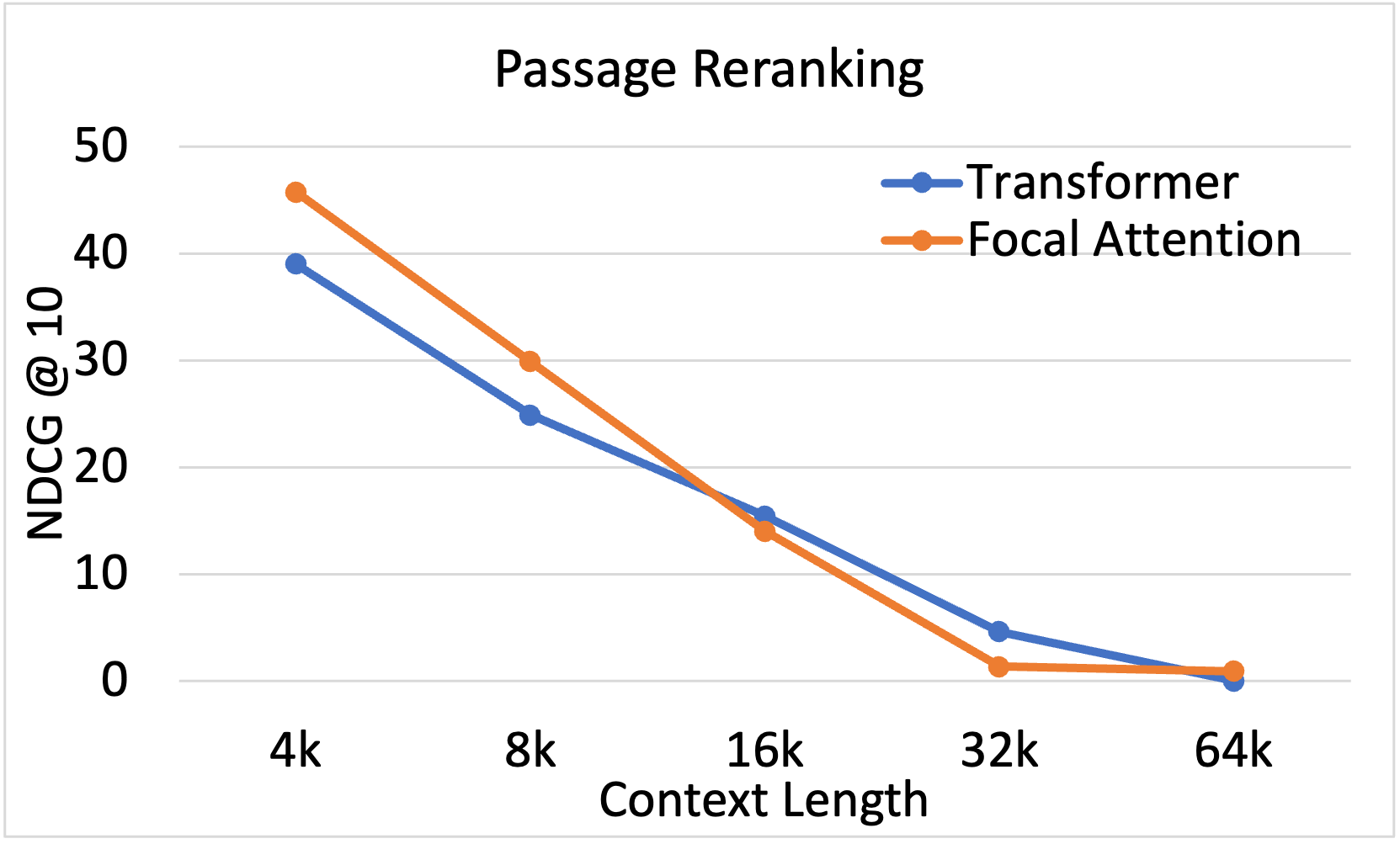}
 \end{subfigure}
 \begin{subfigure}{0.3\linewidth}
 \includegraphics[width=\columnwidth]{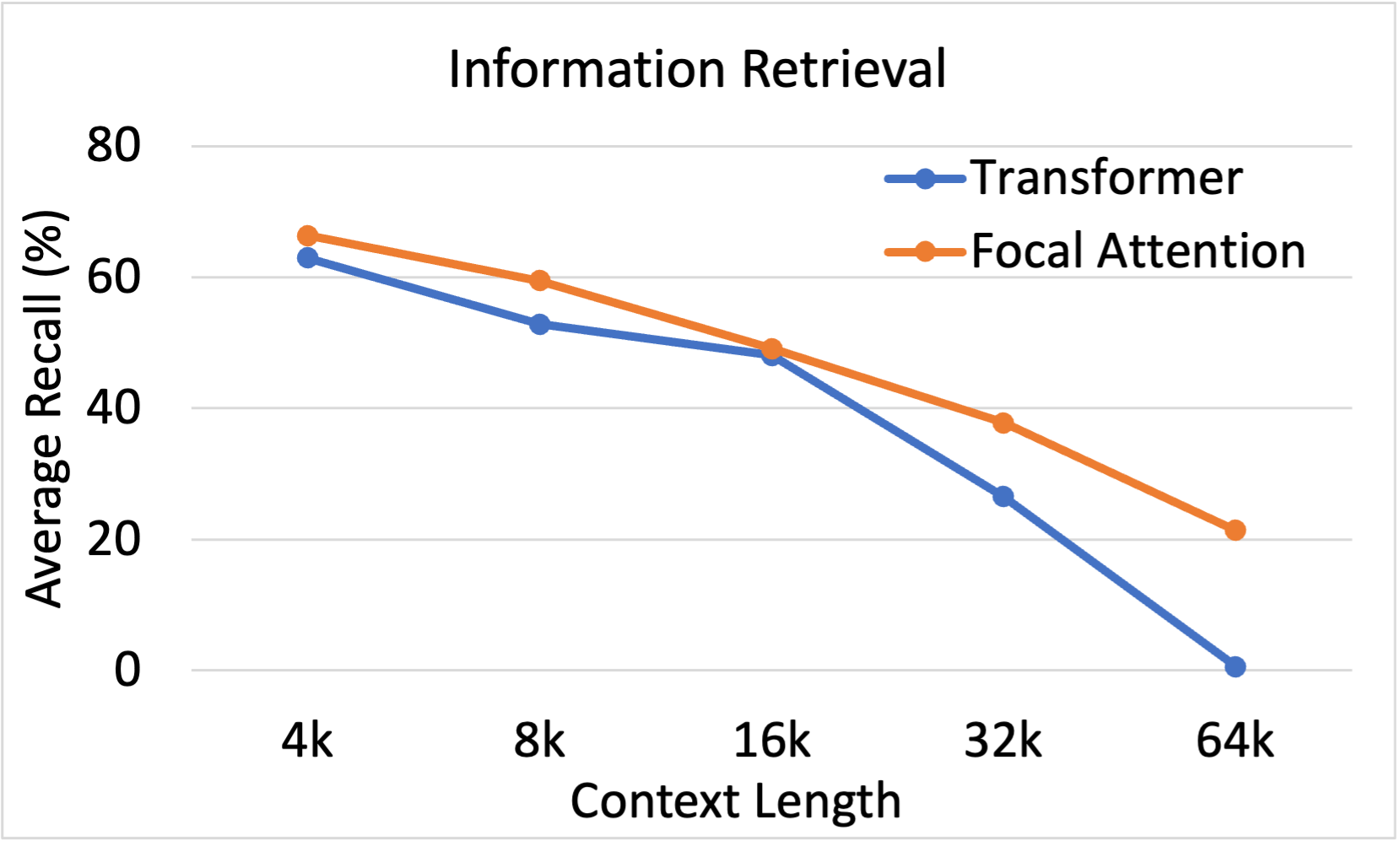}
 \end{subfigure}
\caption{Long Context Capability: \emph{Focal Attention} perform significantly better than transformer for 5 out of 6 task families in HELMET at different context length, each task consists of multiple datasets.}
\label{fig:long-context}
\end{figure*}

\subsection{Long Context Capability} \label{sec:long-context}
%To measure the effectiveness of focal attention at longer context, we consider the 2.7B model trained for 300B tokens with context length 2048. We adapt both models (base transformer and Focal Attention) for 32768 context length by fine tuning it for an additional 5B tokens at that length with longer training samples. We scale up the RoPE $\theta$ from 10K to 500K for this purpose following~\citep{xiong2023effective}. 
We evaluate the long context performance of \emph{Focal Attention} using HELMET framework as discussed in Section~\ref{sec:eval-long}. The evaluations are performed at 5 different context length (from 4K to 64K) for each group of tasks, and the results are presented in Figure~\ref{fig:long-context}. We observe significant performance improvement with \emph{Focal Attention} compared to baseline, with largest gains seen for in-context learning and retrieval tasks. 
Note that the models are fine-tuned at 32K context length, so the 64K context evaluation is performed with dynamic NTK extrapolation following~\citep{peng2023yarn}. We performed this to analyzed the ability of \emph{Focal Attention} to extend the context length beyond the trained value of 32K, and the results in Figure~\ref{fig:long-context} indicate that \emph{Focal Attention} performs significantly better than baseline with extended context length in 4 out of 6 task groups (ICL, Long Document QA, RAG and Retrieval). For Generation with citation task, \emph{Focal Attention} performs better until 16K context, and similar for 32K and 64K context, while the results are mixed for Passage reranking task. We further analyze dataset and task specific performance for a sub-group of tasks in the following sections and extended analysis can be found in Appendix~\ref{sec:apx-retrieval}.

\begin{figure*}
\centering  
 \begin{subfigure}{0.24\linewidth}
\includegraphics[width=\columnwidth]{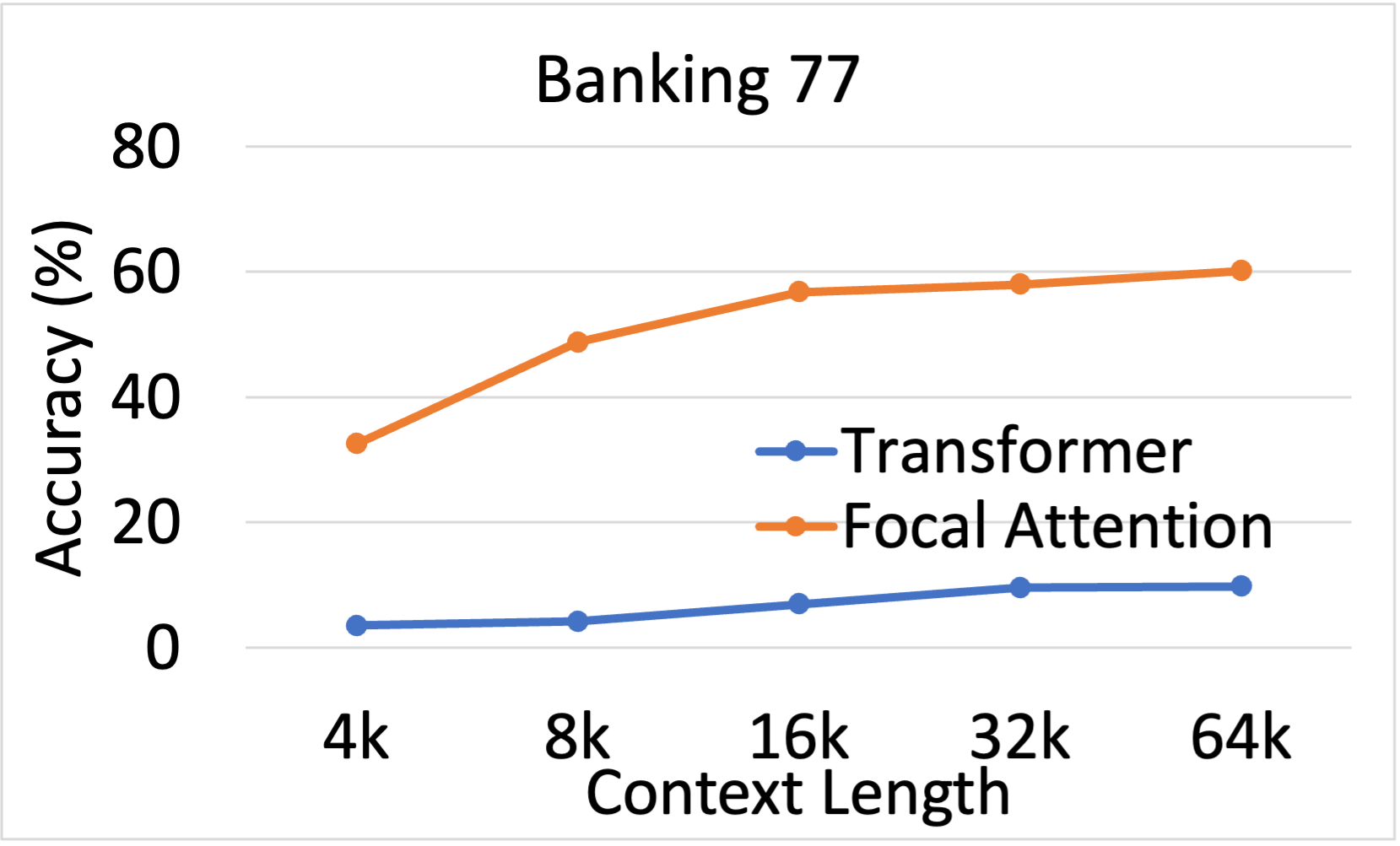}
 \end{subfigure}
 \begin{subfigure}{0.24\linewidth}
 \includegraphics[width=\columnwidth]{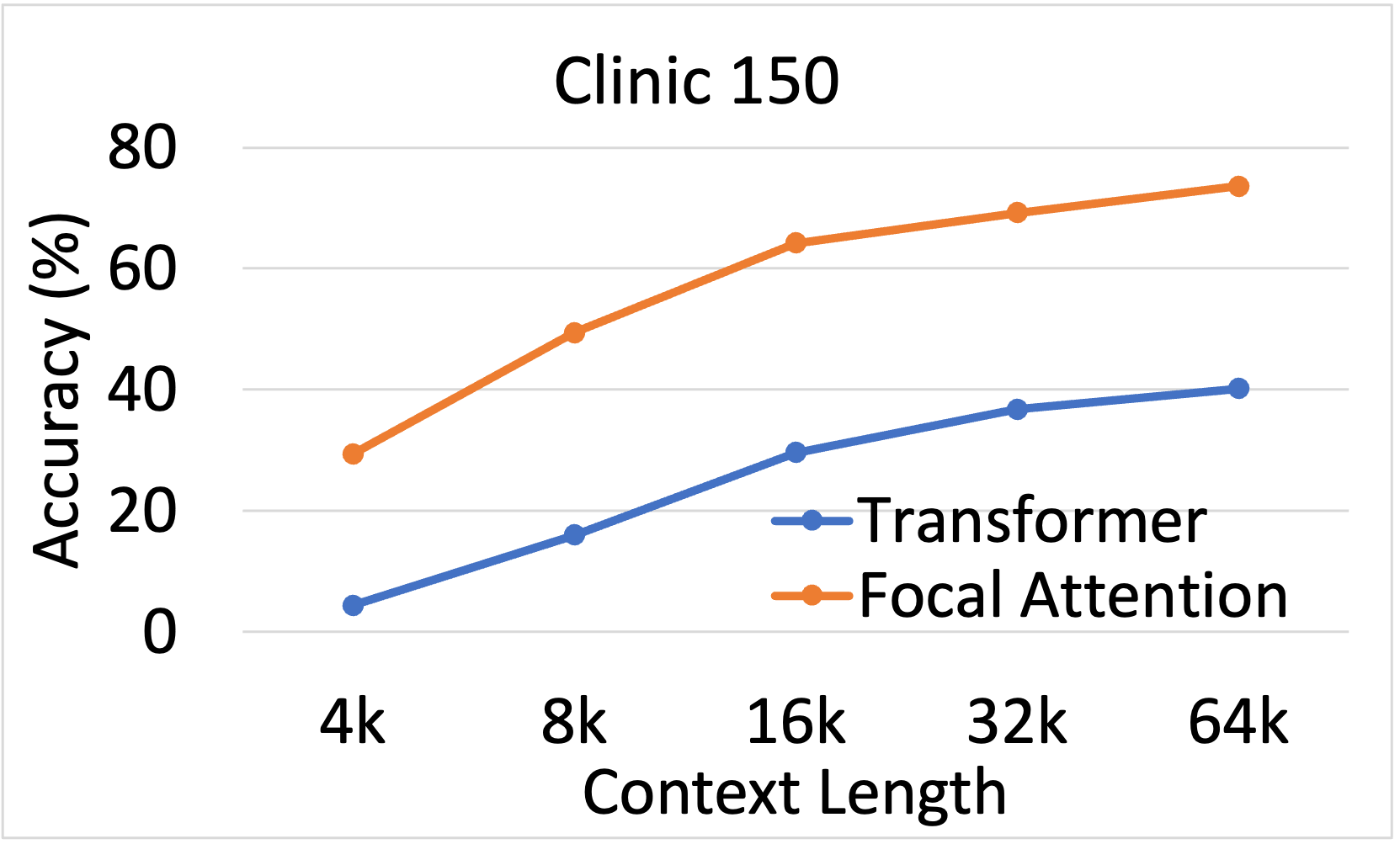}
 \end{subfigure}
% \begin{subfigure}{0.3\linewidth}
% \includegraphics[width=\columnwidth]{icl/trec_coarse.png}
% \end{subfigure}
 \begin{subfigure}{0.24\linewidth}
 \includegraphics[width=\columnwidth]{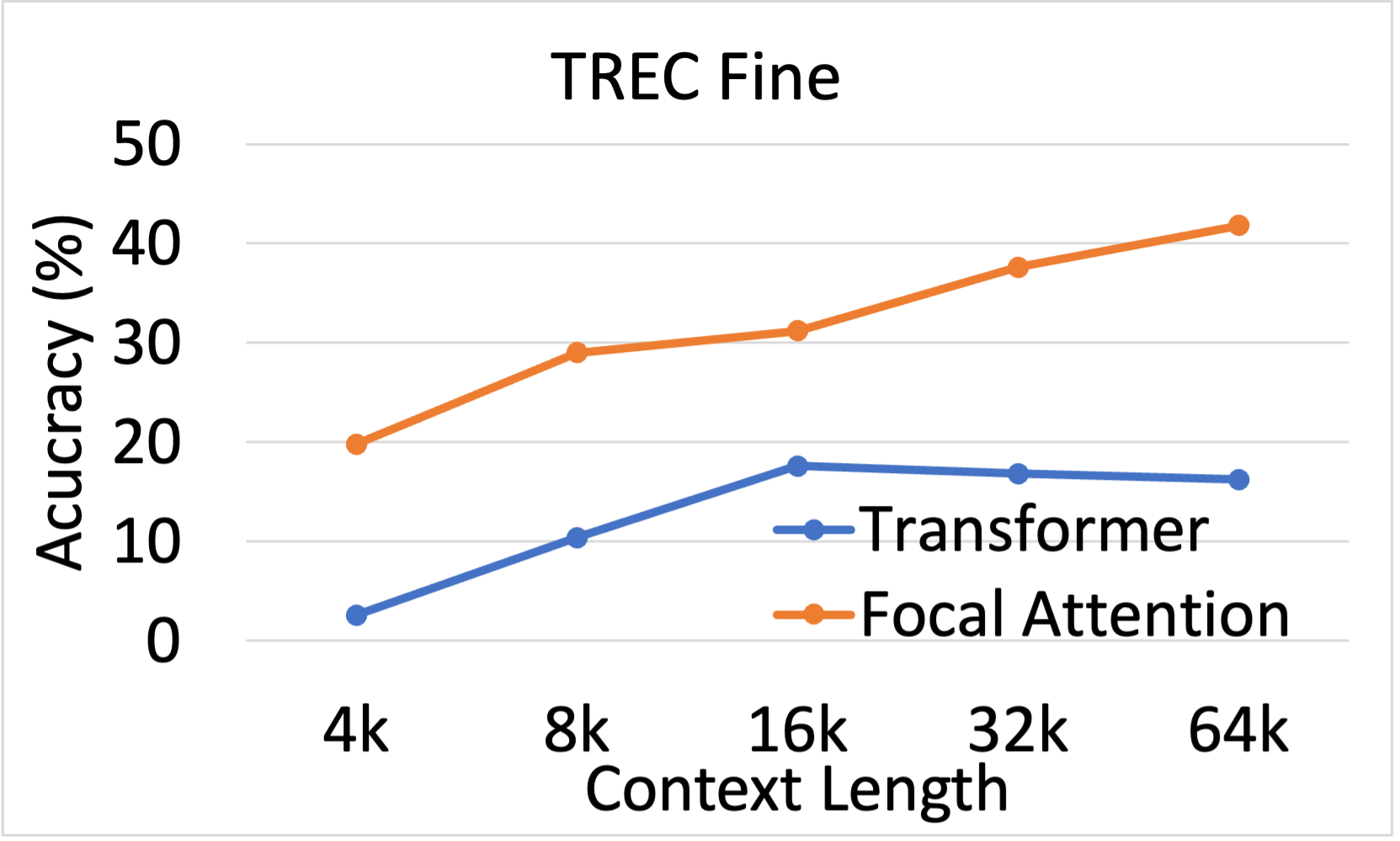}
 \end{subfigure}
 \begin{subfigure}{0.24\linewidth}
 \includegraphics[width=\columnwidth]{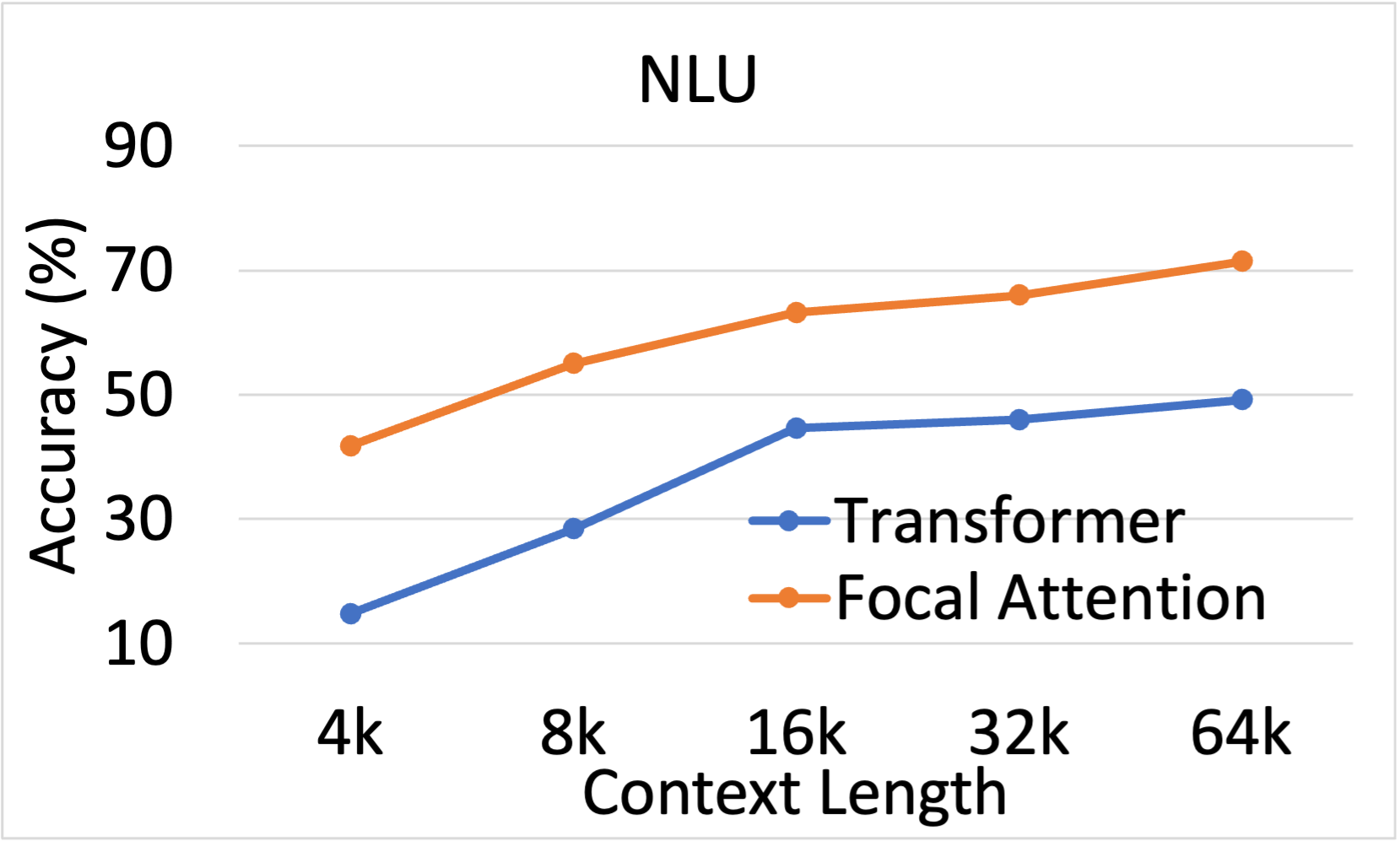}
 \end{subfigure}
\caption{In-Context Learning: \emph{Focal Attention} yields large improvement over transformer for ICL tasks in 4 different datasets for all evaluated context lengths.}
\label{fig:long-icl}
\end{figure*} 

\subsubsection{In-Context Learning}
In-context Learning enables LLMs to adapt to novel tasks by learning from examples provided in the context~\citep{brown2020language}. Long context models enables us to provide hundreds or thousands of examples of a task in the context that helps the model in recognizing the pattern and perform better. In this work we evaluate on datasets with large number of labels: Banking77~\citep{casanueva2020efficient}, Clinic150~\citep{larson2019evaluation}, NLU~\citep{liu2021benchmarking}, TREC Coarse and Fine~\citep{li2002learning}. The number of examples is adjusted according to the context length and the results are presented in Figure~\ref{fig:long-icl}. We observe large improvements in all datasets showing generalizability of our approach. The performance is increasing with more in-context examples indicating room for further improvement. 

\begin{figure*}
\centering  
 \begin{subfigure}{0.24\linewidth}
\includegraphics[width=\columnwidth]{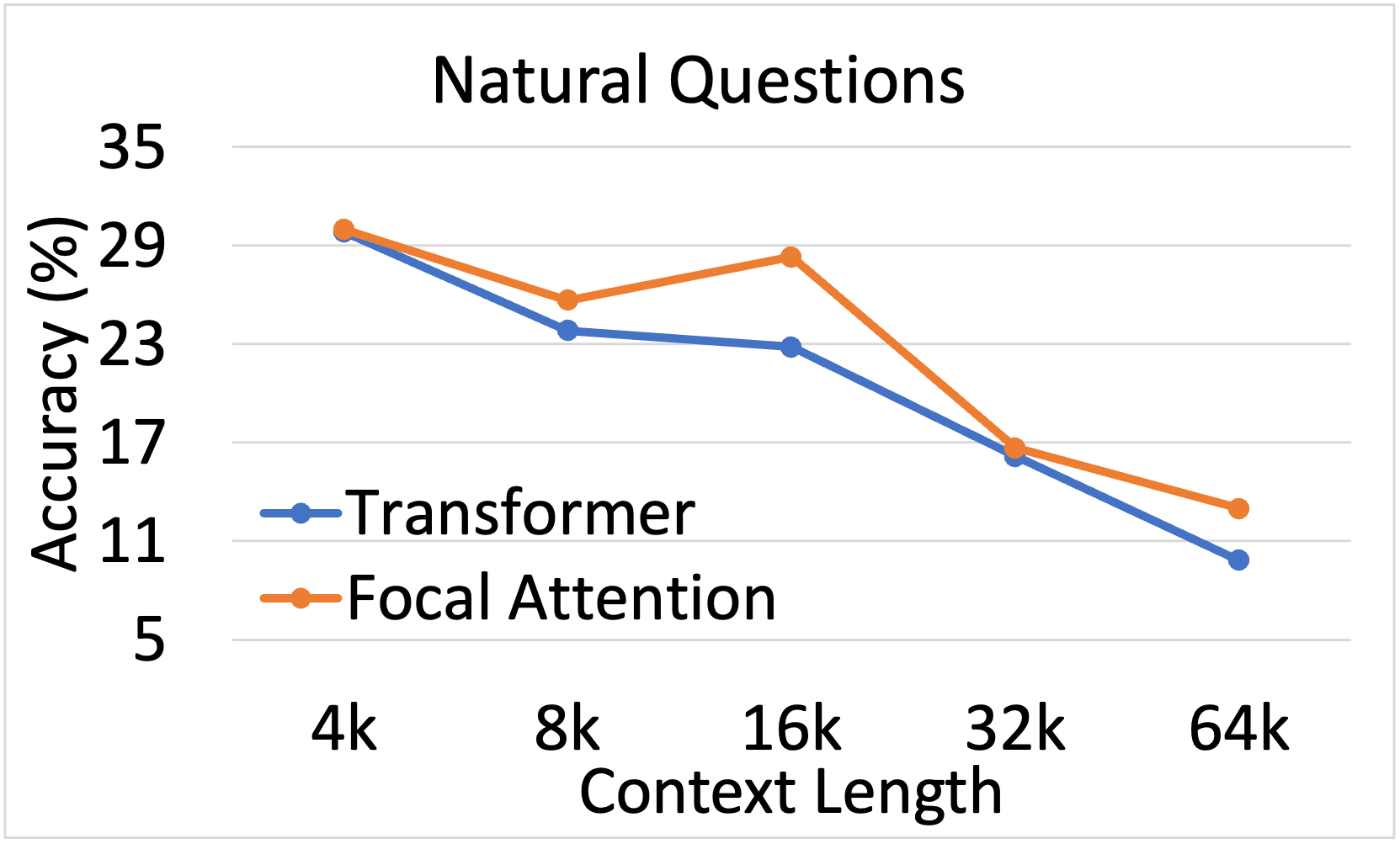}
 \end{subfigure}
 \begin{subfigure}{0.24\linewidth}
 \includegraphics[width=\columnwidth]{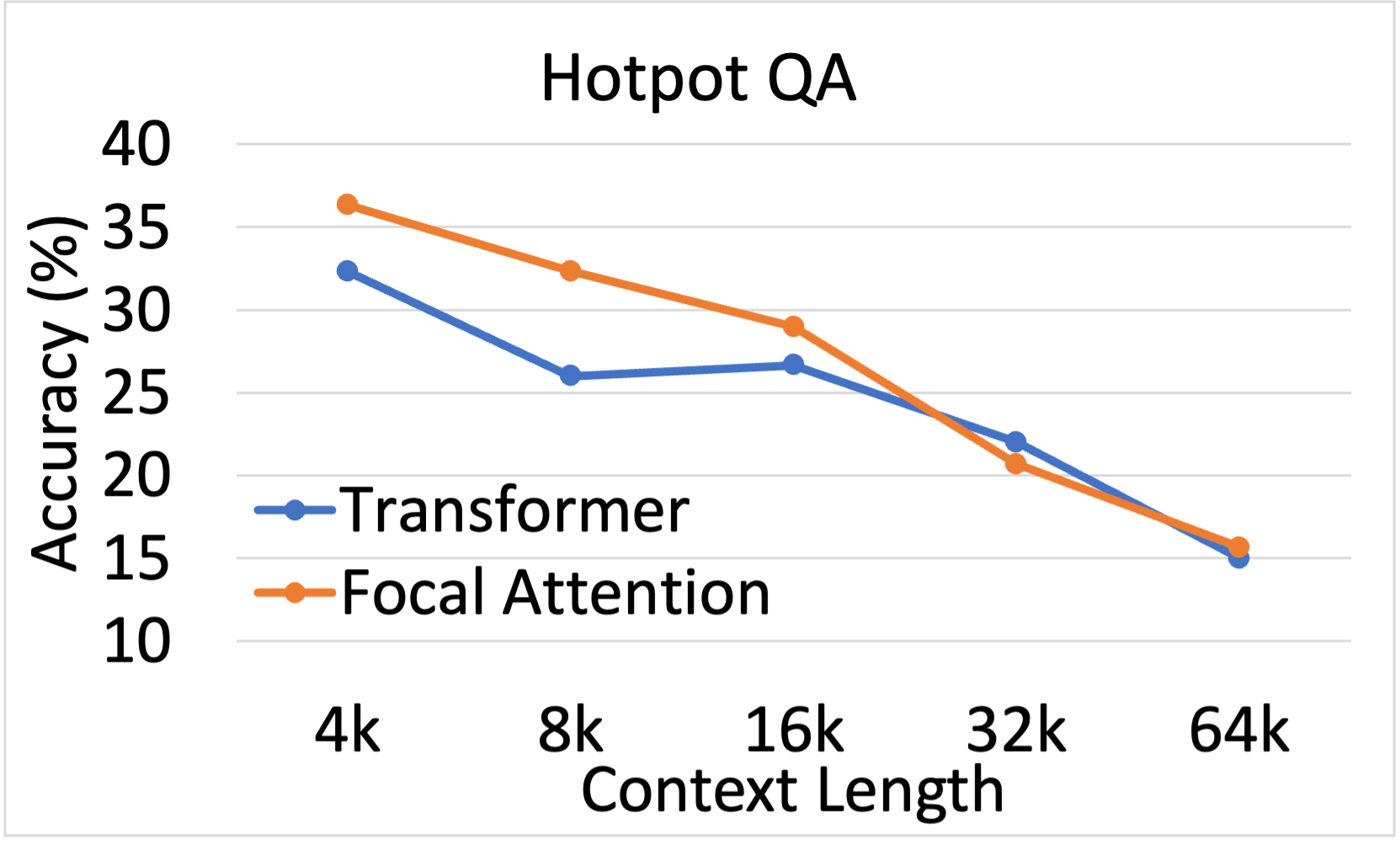}
 \end{subfigure}
 \begin{subfigure}{0.24\linewidth}
 \includegraphics[width=\columnwidth]{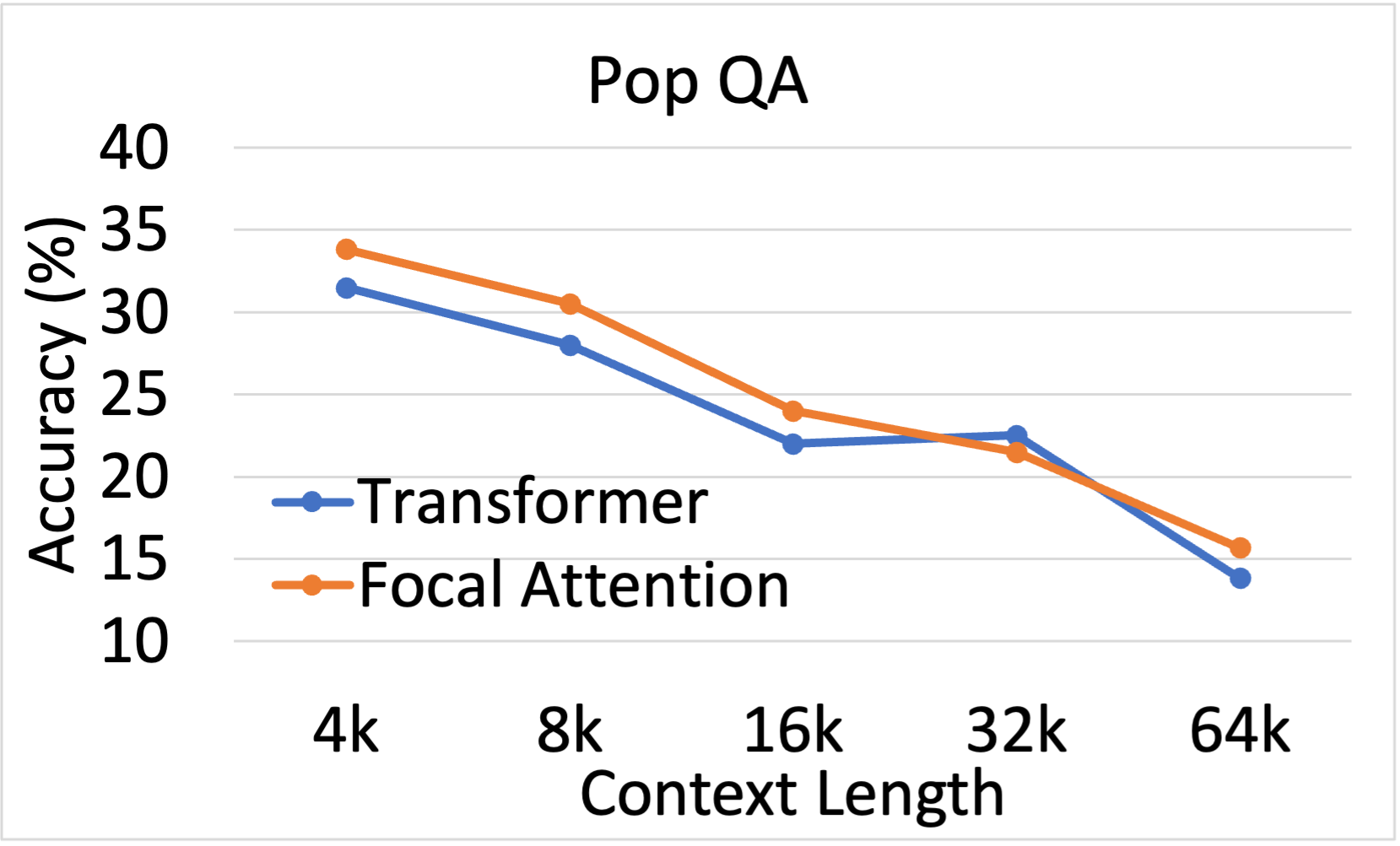}
 \end{subfigure}
 \begin{subfigure}{0.24\linewidth}
 \includegraphics[width=\columnwidth]{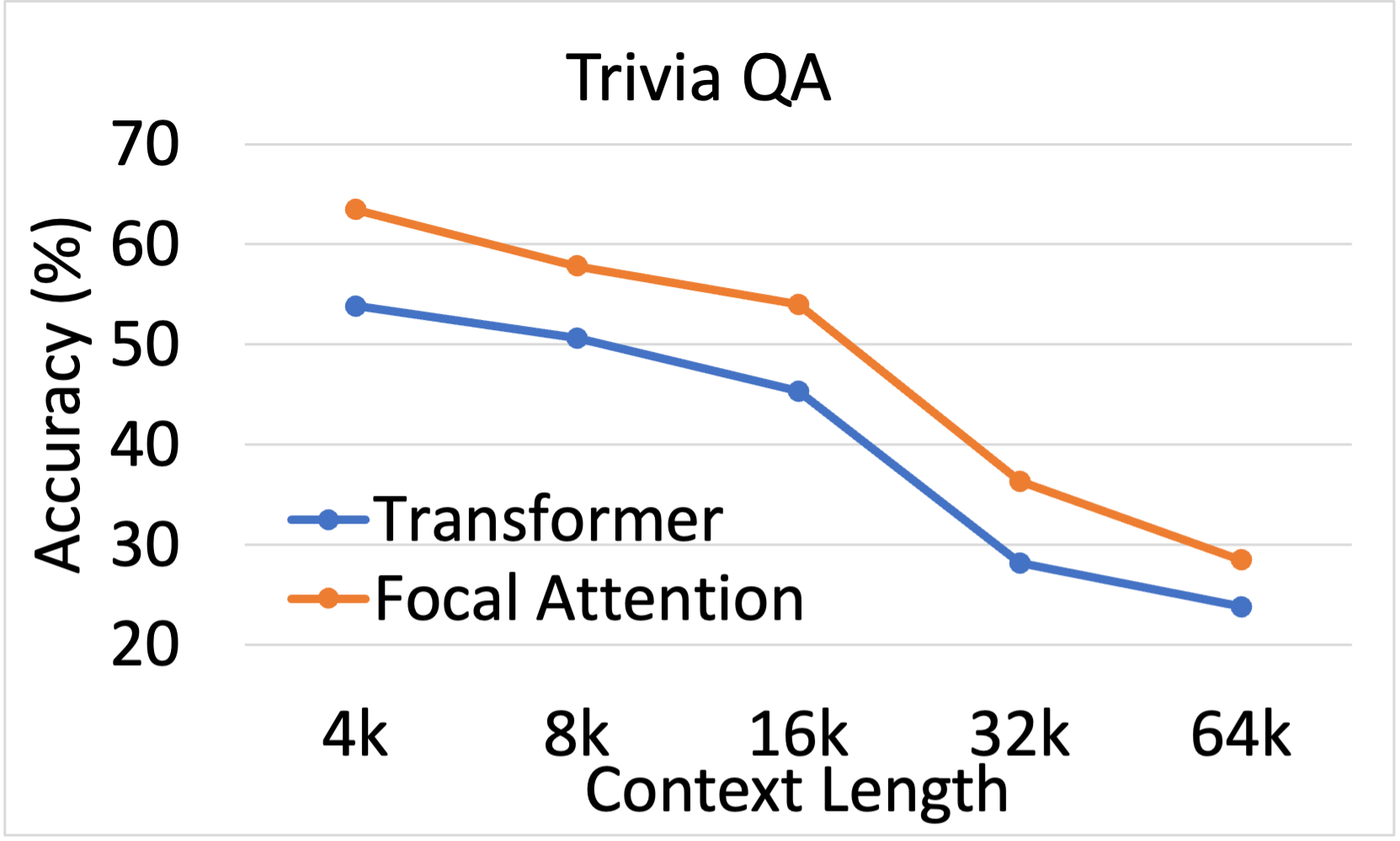}
 \end{subfigure}
\caption{Retrieval Augmented Generation: Performance of \emph{Focal Attention} and transformer model in 4 different open domain question answering task. \emph{Focal Attention} consistently performs better.}
\label{fig:long-rag}
\end{figure*} 
\subsubsection{Retrieval Augmented Generation (RAG)}
Open domain question answering is used as a representative for RAG evaluation where passages from a knowledge corpus is augmented into the context. This context was created by mixing the gold passage with distractor passages (passages that don't contain the answer) that were retrieved from the same corpus instead of random passages~\citep{yen2025helmet}. We evaluate our models on Natural Questions~\citep{kwiatkowski2019natural}, HotpotQA~\citep{yang2018hotpotqa}, PopQA~\citep{mallen2022not} and TriviaQA~\citep{joshi2017triviaqa} and present the results in Figure~\ref{fig:long-rag}. We observe \emph{Focal Attention} performs significantly better (except at 32K for HotpotQA and PopQA) showing its effectiveness for RAG. 

\begin{table*}
\centering
\resizebox{0.95\textwidth}{!}{%
\setlength{\tabcolsep}{4pt}
\begin{tabular}{c|cccccccc}
\toprule
 \textbf{Adapt Temp.} & \textbf{ARC-C} & \textbf{ARC-E} & \textbf{BoolQ} & \textbf{Hellaswag} & \textbf{LAMBADA}  & \textbf{PIQA} & \textbf{Winogrande} & \textbf{Average}\\ 
\midrule 
1 & 31.14 & 55.77 & 62.32 & 58.31 & 51.48 & 73.56 & 55.33 & 55.42 \\
0.4 & 30.55 & {\bf 56.1} & 61.35 & {\bf 59.19} & \textbf{53.85} & \textbf{73.67} & \textbf{56.59} & \textbf{55.9} \\
\midrule 
%\makecell{0.4 (from \\ scratch)} & 32.08 & 60.52 & 63.79 & 62.12 & 53.85 & 74.16 & 59.59 & 58.02 \\
0.4 (from scratch) & 32.08 & 60.52 & 63.79 & 62.12 & 53.85 & 74.16 & 59.59 & 58.02 \\
\bottomrule
\end{tabular}
}
\caption{Comparison of adapting pretrained baseline transformer model to \emph{Focal Attention} with limited data.} 
\label{tab:adapt-perf}
\end{table*}

\begin{figure}
\centering  
 \begin{subfigure}{0.48\columnwidth}
\includegraphics[width=\columnwidth]{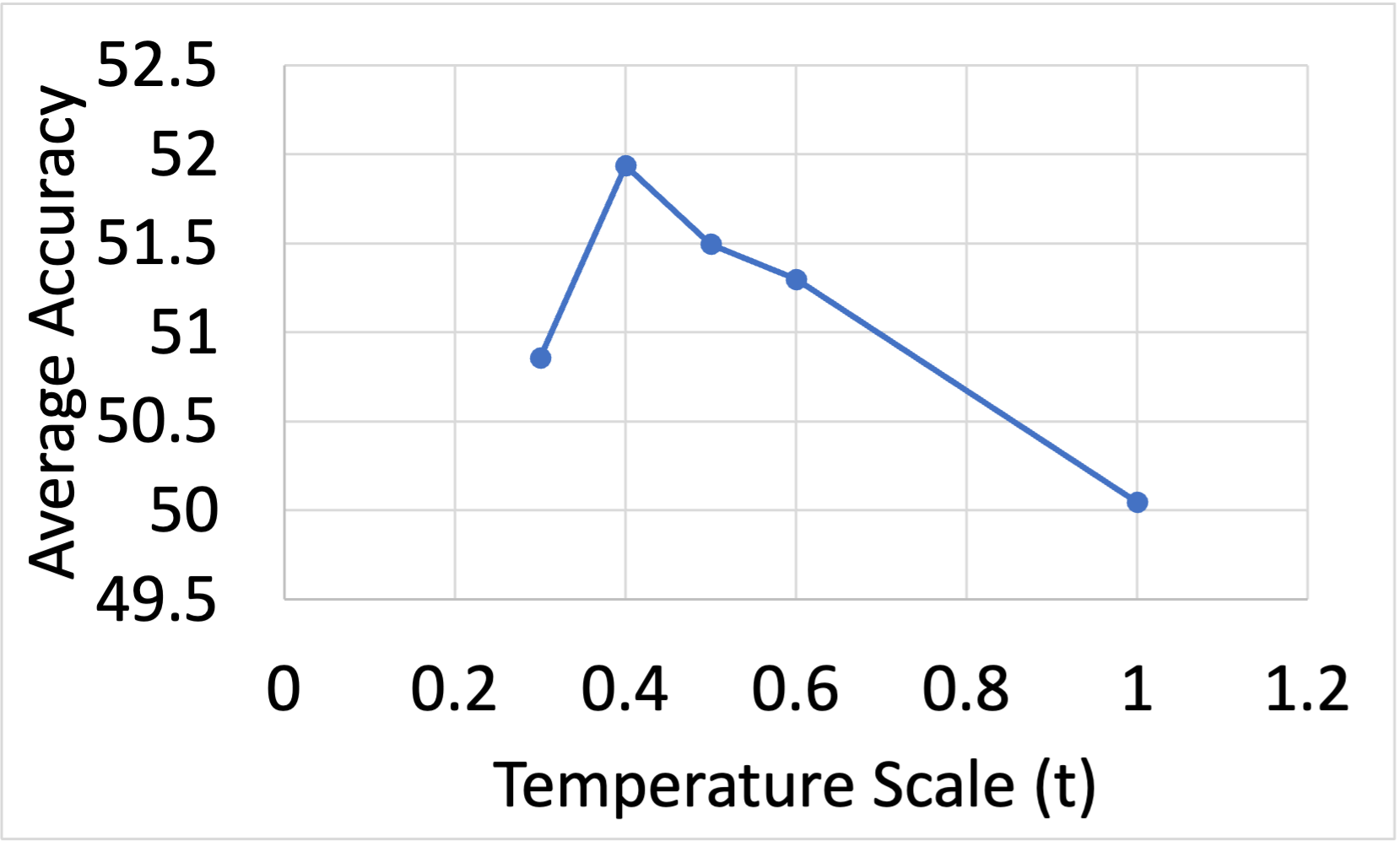}
%\caption{Performance variation with Focal Attention temperature scale.}
\caption{Performance w.r.t. Focal temperature scale.}
\label{fig:vary-temp-scale}
 \end{subfigure}
%\hfill
%\hspace{0.02\textwidth}
 \begin{subfigure}{0.48\columnwidth}
 \includegraphics[width=\columnwidth]{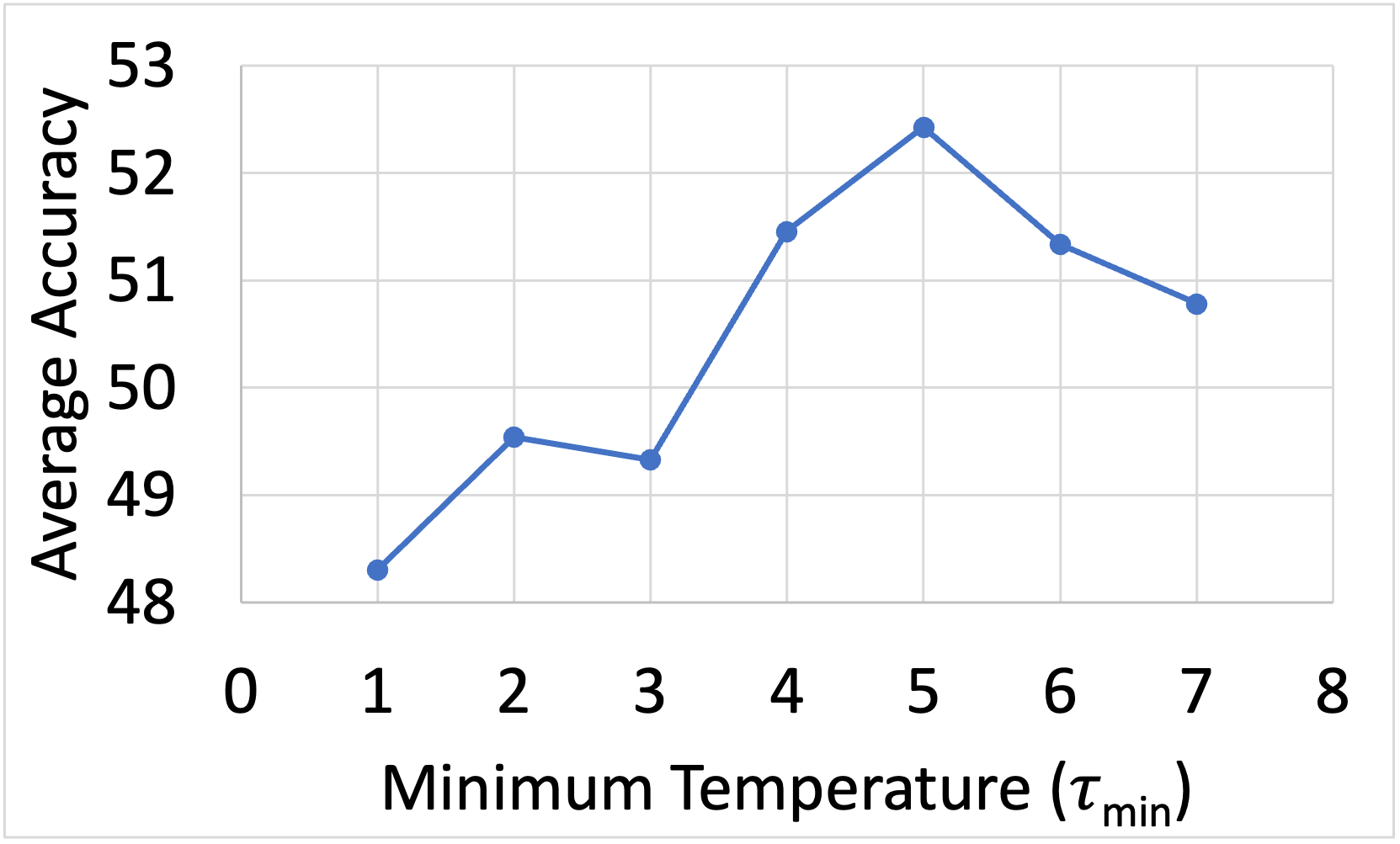}
% \caption{Performance variation with min temperature for learned Focal Attention.}
\caption{Perf. w.r.t. min temp for learned \emph{Focal Attention}.}
\label{fig:vary-min-temp}
 \end{subfigure}
\caption{Effect of \emph{Focal Attention} temperature on downstream task performance.}
\end{figure}

\section{Ablation Studies}
In this section we study the effect of temperature in \emph{Focal Attention} and the possibility of adapting pretrained transformer models to use \emph{Focal Attention}.

\subsection{Effect of Constant Temperature}
We study the effect of varying the temperature scale using the 2.7B parameter model, with 26B total training tokens using the same setup described in Section~\ref{sec:train}. We vary the temperature scale $t$ (in~\ref{eq:focal-const}) from 0.3 - 0.6, with an increment of 0.1. The average downstream task performance on LM-Evaluation-Harness is presented in Figure~\ref{fig:vary-temp-scale}. We observe that for all $t<1$ the model performs better than $t=1$ (corresponding to baseline transformer). The scaling factor $t=0.4$ yield the best performance which we later used to train the 2.7B model for 315B tokens to compare the large tokens training and long context model (Sections~\ref{sec:scale-data} and~\ref{sec:long-context}).

\subsection{Effect of Learned Temperature} \label{sec:effect-learned}
In case of learned temperature, we have two hyperparameter $\tau_{min}$ and $\tau_{max}$ (in~\ref{eq:focal-learn}) to control the attention probability distribution as described in Section~\ref{sec:focal-learn}. We experiment with $\tau_{max}$ = 10 and 11.31 (corresponding to the baseline transformer: sqrt(128) = 11.31). We noticed $\tau_{max}$ = 10 performs marginally better, so chose it for the rest of our experiments. We varied $\tau_{min}$ from 1 to 7 with an increment of 1 and the average task performance is shown in Figure~\ref{fig:vary-min-temp}. We notice that $\tau_{min} = 5$ yields the best performance, and lower values can result in larger degradation indicating that much sharper attention probability distribution can limit the feature selection ability of the model in each layer causing the degradation.  

\subsection{Adapting to Focal Attention} \label{sec:adapt}
Finally, we test the ability of a pretrained baseline transformer model to adapt to \emph{Focal Attention}. We consider the 2.7B model (trained with 315B tokens) and scale the temperature of each attention layer with $t=0.4$ . We continually train the baseline and \emph{Focal Attention} model with learning rate = $5\times10^{-5}$, batch size = 0.26M, for 5K steps (due to limited compute) resulting in 1.3B total training tokens. The downstream task performance of these models are compared in Table~\ref{tab:adapt-perf}. We observe that the adapting to \emph{Focal Attention} improves the performance of transformer model, however its worse than the model trained with $t=0.4$ from scratch. Thus to yield the most benefit of \emph{Focal Attention}, we should use it since the beginning of training. 
%It may be possible to improve the performance of temperature adapted by fine tuning for longer with more tokens, however we did not conduct this experiment due to limited compute. 

%% file: related_work.tex
\section{Related Work}
The attention mechanism in transformer models~\citep{vaswani2017attention} enables each token to weigh the relevance of every other token in the sequence, facilitating contextualized representation learning. Over time, several architectural enhancements have been proposed to improve the scalability and performance of transformers, including efficient attention variants~\citep{beltagy2020longformer, dao2022flashattention, peng2023rwkv} and sparse attention mechanisms~\citep{child2019generating, zaheer2020bigbird}. However, an assumption across these methods is the use of a fixed softmax temperature in computing attention distributions, typically set to the square root of the key/query dimension~\citep{vaswani2017attention}.

%\subsection{Temperature Scaling in Neural Networks}
Temperature scaling has been previously used for calibration~\citep{guo2017calibration} and sample diversity control in autoregressive decoding~\citep{holtzman2019curious}. In the context of language generation, it is commonly employed at the output layer to manipulate entropy of the predicted distribution and encourage either confident or diverse responses~\citep{meister2022typical, lin2022sampling}. However, its application within the core attention mechanism of transformers has remained underexplored. While~\citep{liu2023lost} and~\citep{dong2023raft} highlight that models struggle with locating relevant information in long contexts, they do not explicitly manipulate softmax temperature to address this issue.

%\subsection{Adaptive and Learnable Attention Mechanisms}
Several works have attempted to make attention computation more adaptive. For example, adaptive span~\citep{sukhbaatar2019adaptive} learns dynamic context windows per head, and routing transformers~\citep{roy2021efficient} learn token-to-token routing using clustering. Similarly, parameterizing attention weights or gates has been explored to allow dynamic control~\citep{shen2018disan, zhang2022adaptive}. Our work differs in that it directly modifies the attention softmax operation itself via temperature scaling, offering a simpler means of controlling selectivity.

%\subsection{Attention in Long-Context Models}
Scaling transformers to long contexts has prompted the development of architectures like Longformer~\citep{beltagy2020longformer}, BigBird~\citep{zaheer2020bigbird}, and Mamba~\citep{gu2023mamba}. Other approaches such as Memorizing Transformers~\citep{wu2022memorizing}, Infini-attention~\citep{dai2024infinigen}, and position-interpolation methods~\citep{chen2023extending, chen2023flashformer} aim to increase context length while preserving retrieval capabilities. However, these models often rely on architectural complexity or memory-augmented modules. Our work is orthogonal and complementary: it introduces a simple temperature-controlled mechanism that improves focus and relevance filtering in long sequences without requiring architectural modifications.

%\subsection{Learning to Attend and Interpretability}
%Recent research has explored whether attention distributions align with human intuition or model reasoning~\citep{vig2019analyzing, wiegreffe2019attention}. Some works suggest that sharper attention distributions may improve model interpretability and reduce over-reliance on spurious correlations~\citep{serrano2019attention}. By tuning or learning attention temperature, Focal Attention can potentially enhance the interpretability of model decisions by reducing noise in attention patterns.

%\subsection{Reinforcement Learning and Diverse Sampling}
Reinforcement learning techniques like PPO~\citep{ouyang2022training} and contrastive preference optimization~\citep{yuan2024sequence} have been used to improve LLM alignment and reasoning by filtering diverse generations. These pipelines often depend on diverse decoding via temperature or top-k/p sampling~\citep{lu2022quark, snell2024scaling}. However, they treat sampling temperature as an external mechanism. Our work shows that similar diversity or selectivity can be gained by scaling attention temperature internally, potentially yielding benefits for both training and inference.

%% file: appendix.tex
\section*{Appendix}
\section{Model Architecture} \label{sec:apx-arch}
We train a family of autoregressive Transformer-based language models ranging from 400M to 9.5B parameters. Each model follows a standard decoder-only architecture with rotary position embeddings and pre-layer normalization. Table \ref{tab:arch} details the hyperparameters for each configuration, including hidden size, intermediate size in the feed-forward networks, number of attention heads, and number of Transformer layers. The depth and width of the models scale proportionally with parameter count, following a smooth scaling curve to ensure stable optimization across sizes. The largest model (9.5B) employs 36 layers with a hidden dimension of 4608, while the smallest model (400M) uses 24 layers with a hidden dimension of 1024. All models share the same tokenizer and vocabulary, allowing for direct comparison of performance under controlled scaling.

\begin{table}[h]
\centering
%\resizebox{\columnwidth}{!}{%
\small
\setlength{\tabcolsep}{3pt}
\begin{tabular}{l|cccc} 
\toprule
{\bf \# Params} & \makecell{\bf Hidden \\ \bf Size} & \makecell{\bf Intermediate \\ \bf Size} & {\bf \# Heads} & {\bf \# Layers}  \\ 
\midrule 
400M & 1024 & 3072 & 8 & 24 \\
777M & 1536 & 4096 & 12 & 24 \\
1.3B & 2048 & 5504 & 16 & 24 \\
2.7B & 2560 & 6912 & 20 & 32 \\
6.7B & 4096 & 11008 & 32 & 32 \\
9.5B & 4608 & 12288 & 36 & 36 \\
\bottomrule
\end{tabular}
%}
\caption{Detailed architecture for models with Parameters 400M - 9.5B, used in this work.}
\label{tab:arch}
\end{table}

\section{Training Hyperparameters} \label{sec:apx-hyp}
All models are trained for 100K steps using a global batch size of 0.26M tokens. The learning rate is scaled inversely with model size, consistent with empirical findings from prior scaling studies \citep{kaplan2020scaling,hoffmann2022training}, to stabilize optimization for larger models while enabling faster convergence for smaller ones. We employ a cosine learning rate decay schedule with a short warm-up period of 2K steps, after which the learning rate decreases smoothly toward zero.

Table \ref{tab:hyperparameter} summarizes the exact configurations for each model size. The smallest model (400M) uses a peak learning rate of $4\times 10^{-3}$, while the largest models (6.7B and 9.5B) use $4\times 10^{-3}$. All other optimizer and regularization settings (e.g., AdamW parameters, weight decay, dropout rates) are held constant across sizes to isolate the effect of scale and learning rate. This consistent setup allows for direct comparison of scaling behavior while minimizing confounding factors.
$4\times 10^{-3}$
\begin{table}[h]
\centering
\small
\setlength{\tabcolsep}{6pt}
\begin{tabular}{l|ccc} 
\toprule
\makecell{\textbf{Model}} & \makecell{\textbf{Batch} \\ \textbf{Size}} & \makecell{\textbf{Learning} \\ \textbf{Rate}} & \makecell{\textbf{\# Training} \\ \textbf{Steps}} \\
\midrule 
400M & 0.26M & $4\times 10^{-3}$ & 100K \\
777M & 0.26M & $2\times 10^{-3}$ & 100K \\
1.3B & 0.26M & $1\times 10^{-3}$ & 100K \\
2.7B & 0.26M & $5\times 10^{-4}$ & 100K \\
6.7B & 0.26M & $4\times 10^{-4}$ & 100K \\
9.5B & 0.26M & $4\times 10^{-4}$ & 100K \\
\bottomrule
\end{tabular}
\caption{Training hyperparameters for different sized models.}
\label{tab:hyperparameter}
\end{table}

\section{Long Context Evaluation} \label{sec:apx-eval-long}
We evaluate the long context capability of our models with HELMET~\citep{yen2025helmet} framework. The tasks are categorized in 6 groups, with each group consisting of multiple datasets: (a) Many-shot in-context learning (ICL): Evaluates a model's ability to learn new tasks from a long prompt containing multiple examples, highlights if the model can retain and generalize over extended sequences, (b) Long Document Question Answering (LongQA): Measures a model's ability to locate and synthesize relevant information across lengthy documents to answer questions accurately, (c) Retrieval-Augmented Generation (RAG): Evaluates an LLM's ability to incorporate retrieved, long-context information into its responses, testing grounding and relevance under extended contexts, (d) Passage Reranking: Measures the model’s ability to score and order relevant passages from a large context, crucial for downstream tasks like open-domain QA, (e) Generation with Citations: Evaluates the model’s ability to generate factual outputs grounded in long context inputs while providing accurate source attributions, and (f) Synthetic Recall: Tests whether a model can retrieve exact facts or tokens inserted into long synthetic contexts, probing precise memory over long ranges.

%Retrieval-Augmented Generation (RAG): Evaluates an LLM's ability to incorporate retrieved, long-context information into its responses, testing grounding and relevance under extended contexts.
%In-Context Learning (ICL): Assesses how well a model learns patterns from a long prompt containing multiple examples, highlighting its ability to retain and generalize over extended sequences.

%Long Document Question Answering: Measures a model's ability to locate and synthesize relevant information across lengthy documents to answer questions accurately.

%Synthetic Recall: Tests whether a model can retrieve exact facts or tokens inserted into long synthetic contexts, probing precise memory over long ranges.

%Generation with Citations: Evaluates the model’s ability to generate factual outputs grounded in long context inputs while providing accurate source attributions.

%Passage Reranking: Measures the model’s ability to score and order relevant passages from a large context, crucial for downstream tasks like open-domain QA.

\begin{figure*}
\centering  
 \begin{subfigure}{0.4\linewidth}
\includegraphics[width=\columnwidth]{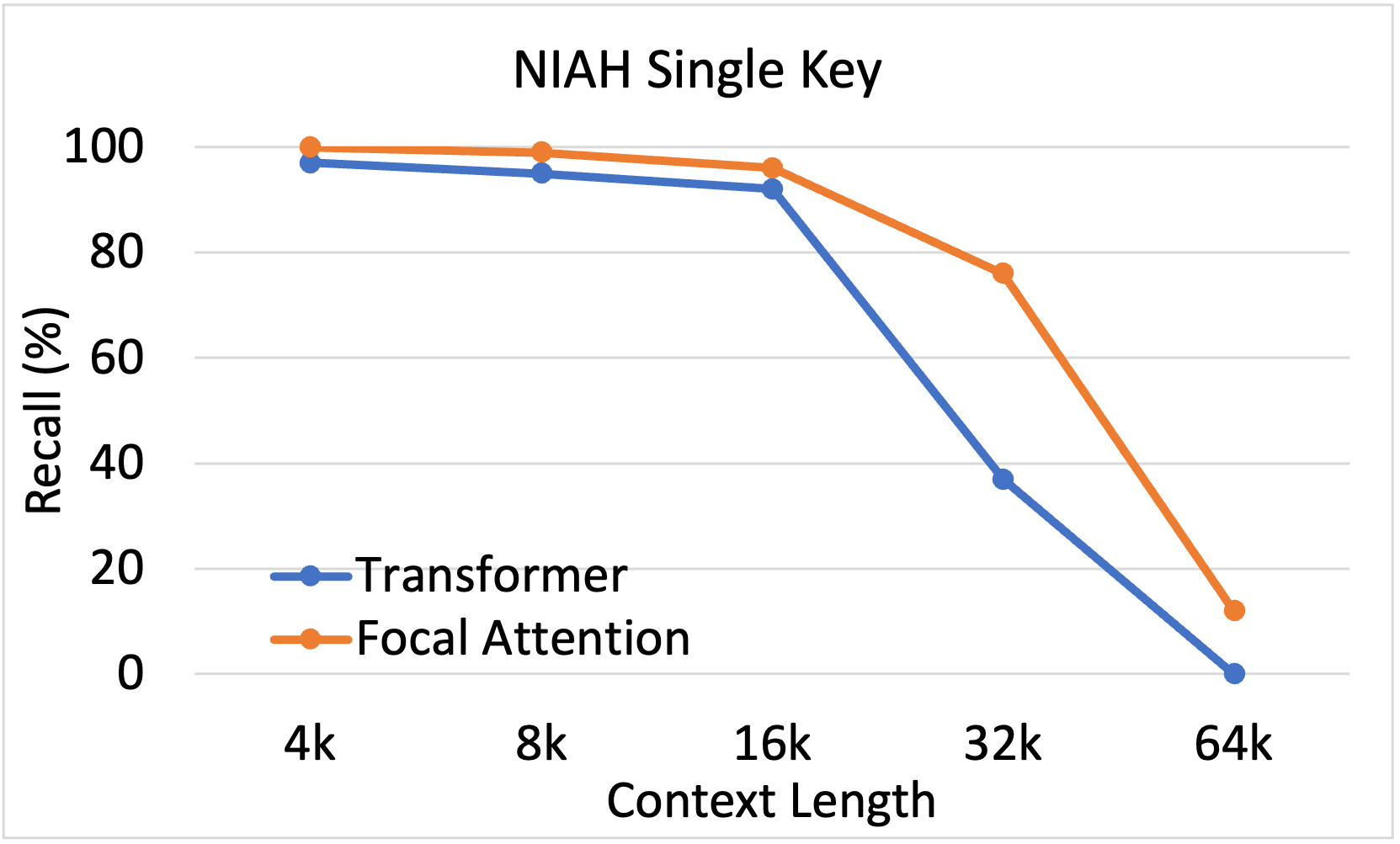}
 \end{subfigure}
 \begin{subfigure}{0.4\linewidth}
 \includegraphics[width=\columnwidth]{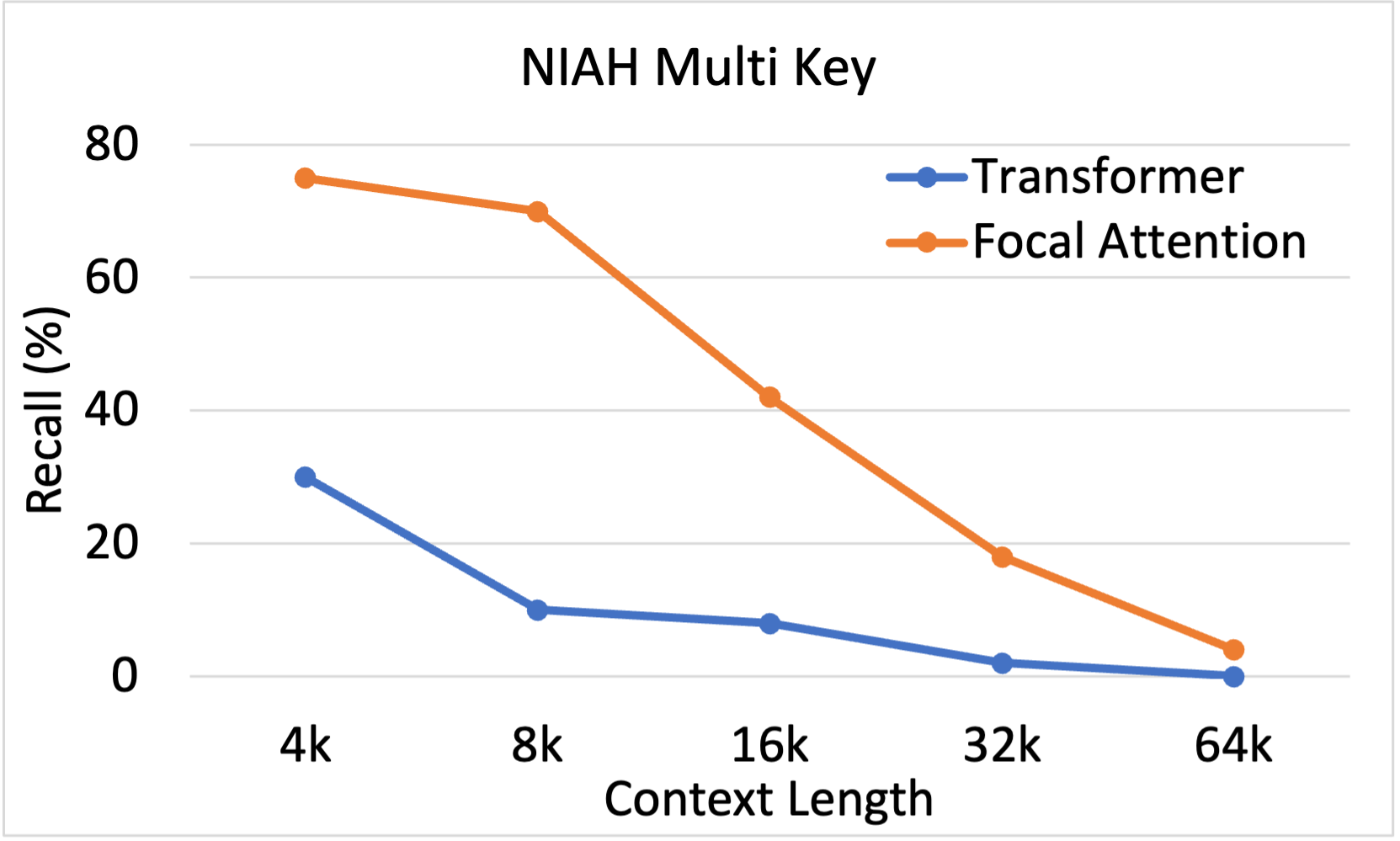}
 \end{subfigure}
 \begin{subfigure}{0.4\linewidth}
 \includegraphics[width=\columnwidth]{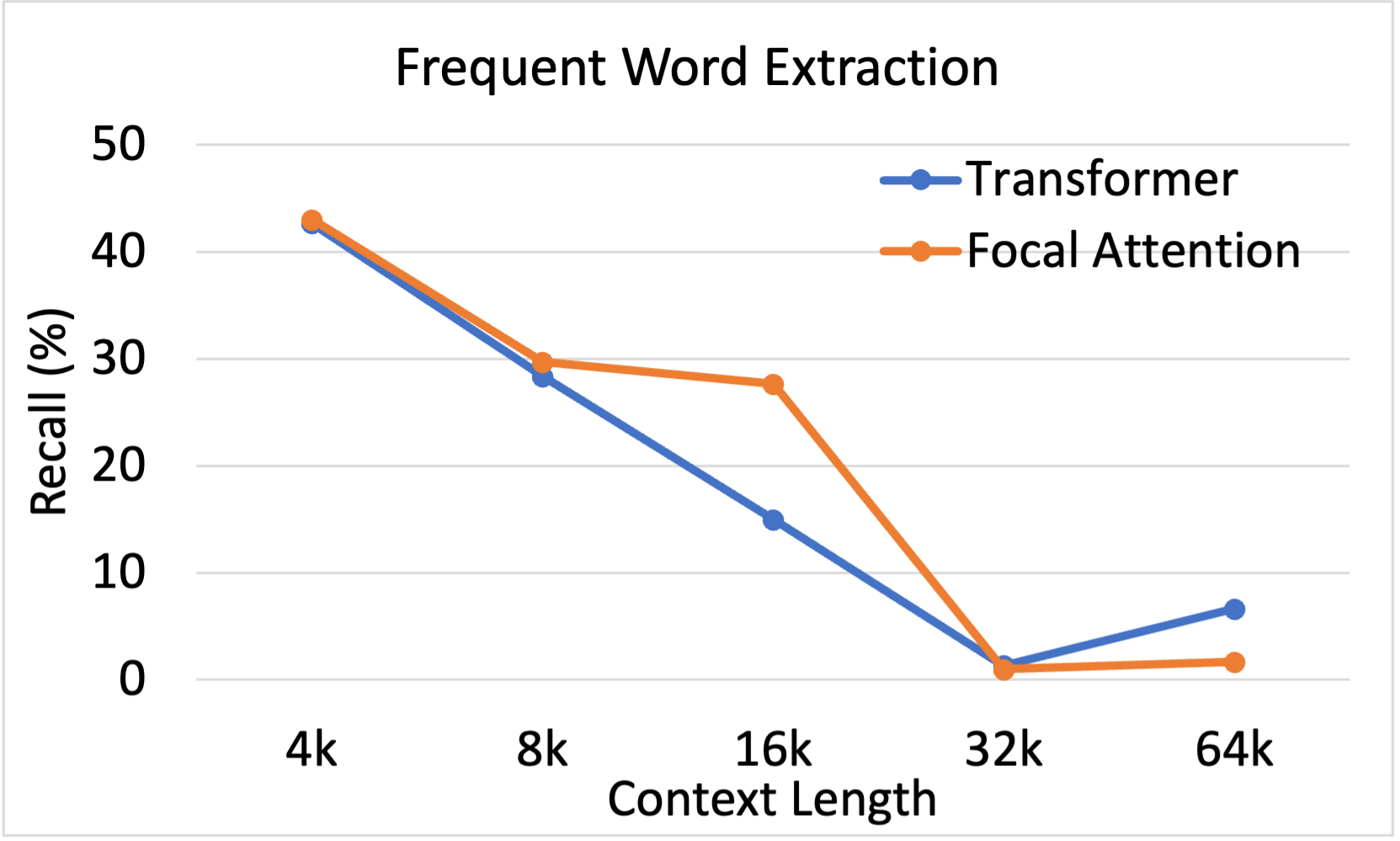}
 \end{subfigure}
 \begin{subfigure}{0.4\linewidth}
 \includegraphics[width=\columnwidth]{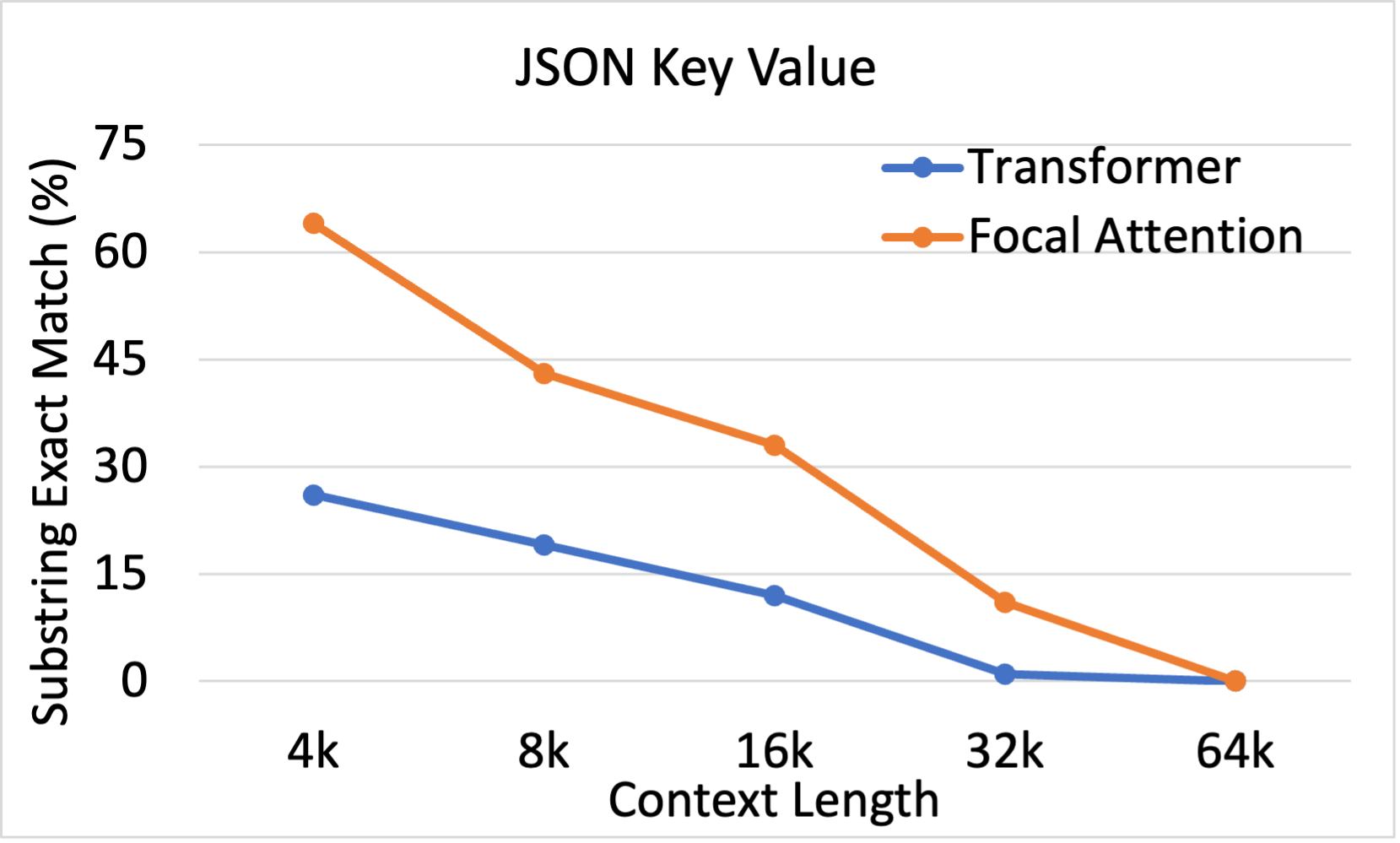}
 \end{subfigure}
\caption{Information Retrieval: Comparing Focal Attention with transformer for 4 retrieval tasks with different difficulty. In each case, Focal Attention performs similar or better for increasing context lengths.}
\label{fig:long-recall}
\end{figure*} 

\section{Information Retrieval} \label{sec:apx-retrieval}
Needle-in-a-haystack~\citep{niah2023} is a popular task to test the ability of LLMs to retrieve key information (i.e. needle) hidden in a long distractor text (i.e. haystack). This simple test was later extended to more challenging variants in RULER benchmark~\citep{hsieh2024ruler} which can be closer to real world tasks. Here, we select a subset of these tasks to compare Focal attention with transformer: (i) single key of type UUID (relatively difficult) hidden in the context of an essay, (ii) multiple needle inserted into an essay and one needle needs to be retrieved. extra needles work as a distractor, making it more challenging, (iii) for the frequent word extraction, the model needs to return top-k most frequent words in the context, and (iv) from a list of JSON-formatted key-value pairs, the model needs to return the value associated with a
specific key~\citep{liu2023lost}. We compare Focal attention with transformer and present the results in Figure~\ref{fig:long-recall}. Focal attention performs significantly better with large improvement in 3 out of 4 tasks.